\documentclass{article}

    \PassOptionsToPackage{numbers, compress}{natbib}
 \usepackage[preprint]{neurips_2026}

\usepackage[utf8]{inputenc} 
\usepackage[T1]{fontenc}    
\usepackage{hyperref}       
\usepackage{url}            
\usepackage{booktabs}       
\usepackage{amsfonts}       
\usepackage{nicefrac}       
\usepackage{microtype}      
\usepackage{xcolor}         
\usepackage{caption}
\usepackage{subcaption}
\usepackage{microtype}
\usepackage{hyperref}
\usepackage{url}
\usepackage{booktabs}
\usepackage{graphicx}   
\usepackage{float}      
\usepackage{booktabs}   
\usepackage{makecell}   
\usepackage{tcolorbox}  
\tcbuselibrary{breakable}
\usepackage{listings}
\usepackage{xcolor}
\newtcolorbox{promptbox}[1][]{
  colback=gray!10!white, 
  colframe=gray!60!black, 
  title=\textbf{System Prompt},
  fonttitle=\bfseries,
  rounded corners,
  boxrule=0.5mm,
  breakable,
  #1
}
\usepackage[T1]{fontenc}
\usepackage{wrapfig}

\title{Unlocking LLM Creativity in Science through Analogical Reasoning}

\author{Andrew Shen \\
Stanford University\\
\texttt{ashen7@stanford.edu} \\
\And
Shaul Druckmann \\
Stanford University\\
\texttt{shauld@stanford.edu} \\
\And
James Zou \\
Stanford University\\
\texttt{jamesz@stanford.edu} \\
}

\begin{document}

\maketitle

\begin{abstract}

Autonomous science promises to augment scientific discovery, particularly in complex fields like biomedicine. However, this requires AI systems that can consistently generate novel and diverse solutions to open-ended problems. We evaluate LLMs on the task of open-ended solution generation and quantify their tendency to mode collapse into low-diversity generations. To mitigate this mode collapse, we introduce analogical reasoning (AR) as a new approach to solution generation. AR generates analogies to cross-domain problems based on shared relational structure, then uses those analogies to search for novel solutions. Compared to baselines, AR discovers significantly more diverse generations (improving solution diversity metrics by 90-173\%), generates novel solutions over 50\% of the time (compared to as little as 1.6\% for baselines), and produces high-quality analogies. To validate the real-world feasibility of AR, we implement AR-generated solutions across four biomedical problems, yielding consistent quantitative gains. AR-generated approaches achieve a nearly 13-fold improvement on distributional metrics for perturbation effect prediction, outperform all baselines on AUPRC when predicting cell-cell communication, infer brain region interactions with a high Spearman correlation ($\rho$=0.729) to published methods, and establish state-of-the-art performance on 2 datasets for oligonucleotide property prediction. The novel and diverse solutions produced by AR can be used to augment the search space of existing solution generation methods.

\begin{figure}[htbp]
    \centering
    \includegraphics[scale=0.56]{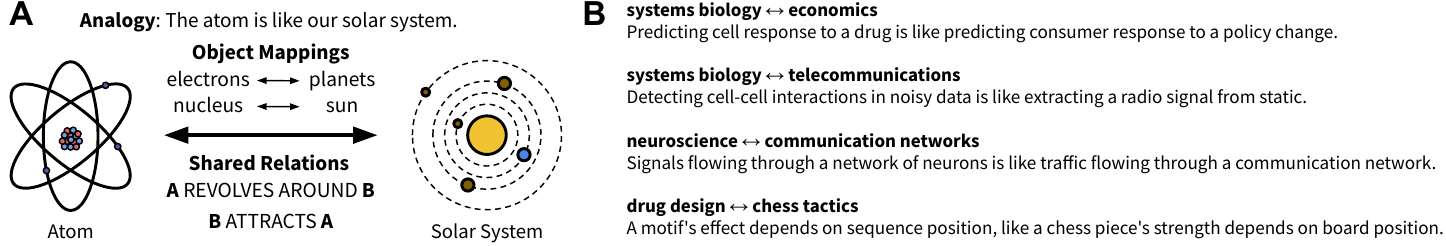}
    
    \vspace{0.25em} 
    
    \includegraphics[scale=0.335]{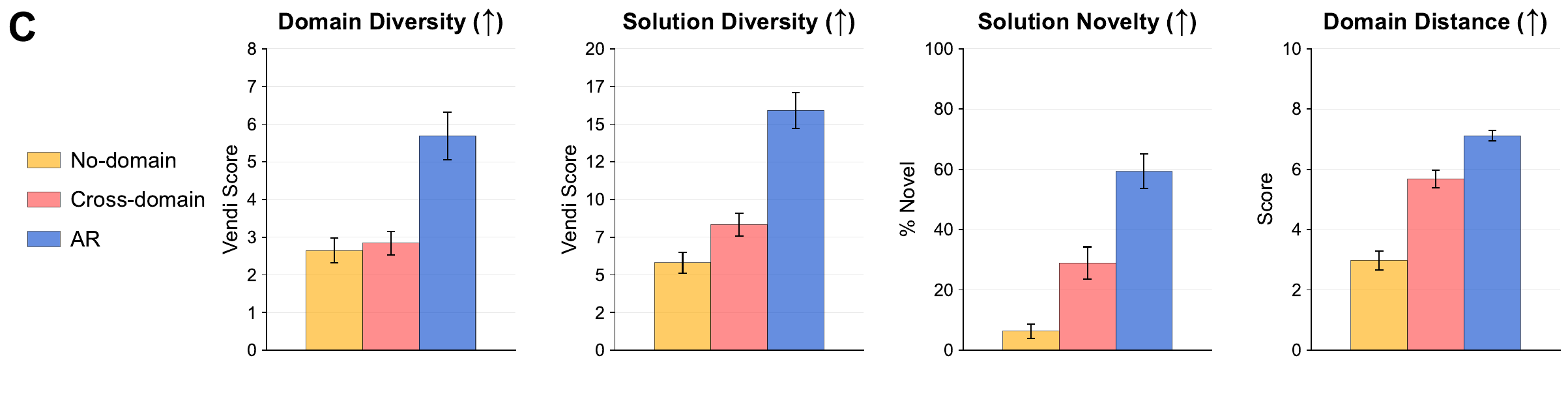}
    
    \caption{(A) An analogy is composed of object mappings and shared relations. Object mappings denote equivalent objects across domains. The relations between object A (electrons or planets) and object B (nucleus or sun) are shared between both domains \citep{structure-mapping, atom_solarsystem}. (B) Example analogies found by the analogical reasoning system. (C) Analogical reasoning (AR) performance against baselines aggregated across all LLMs on the 3 evaluation axes: generation diversity, solution novelty, analogy quality. Generation diversity is measured using Vendi Score, and a higher Vendi Score denotes more diverse generations \citep{vendi}. Domain distance denotes distance between problem and analogous domains.}
    \label{fig:overview}
\end{figure}
\end{abstract}

\section{Introduction}

Automated scientific discovery has emerged as a promising area of research. Frameworks like the AI Scientist \citep{sakana-ai-scientist} demonstrate that AI systems can execute the entire scientific pipeline, from ideation to manuscript generation. In biomedicine, where there exist many unsolved problems, various approaches have been proposed to accelerate scientific breakthroughs. The Virtual Lab simulates a real-life lab environment through collaborating cross-domain LLM agents, resulting in the design of novel COVID-19 binders \citep{virtual-lab}. Kosmos performs long-horizon iteration that discovered novel mechanisms for aging \citep{kosmos}. The AI Co-scientist applies tournament evolution to hypothesis generation, resulting in promising new disease targets \citep{towards-an-ai-coscientist}. These efforts demonstrate AI's immense potential to augment the scientific process. However, the success of autonomous science relies on the ability of AI systems to consistently generate novel and diverse approaches to research problems. Proposed solutions must be novel to drive research progress past existing work. Furthermore, even the most promising solutions may be invalidated after empirical testing. An AI system capable of generating diverse candidate solutions would allow researchers to tackle problems from many angles.

In this work, we evaluate LLMs on open-ended solution generation to assess the novelty and diversity of their outputs. The task of open-ended solution generation consists of generating candidate solutions to solve a given research problem. On this task, we find that current LLMs collapse to semantically similar generations, which is a problematic trend if LLMs are to serve as the engines for autonomous science. To the best of our knowledge, we are the first to systematically quantify the phenomenon of mode collapse in open-ended solution generation for research problems. To mitigate this mode collapse, we introduce analogical reasoning (AR) as a new approach to solution generation. AR uses shared relational structure to transfer knowledge between disparate domains. We use AR for solution generation by prompting LLMs to generate cross-domain analogies, which we then use to search for solutions. We find that solutions generated by AR are significantly more diverse than those from baseline methods. AR-proposed solutions are also more novel, and the underlying analogies are high quality. We confirm this result in a human study, where AR solutions were consistently scored as more novel than baseline solutions. Furthermore, we evaluate our LLM-judged novelty and analogy quality scores with human annotations and find they align closely with expert judgment, achieving correlation and agreement rates comparable to human-to-human baselines. To validate the real-world feasibility of AR, we implement AR-generated solutions across four biomedical case studies. On the tasks of perturbation effect prediction, cell-cell communication inference, brain region interaction inference, and oligonucleotide sequence property prediction, AR-generated solutions exhibit consistent quantitative gains over established baselines. We establish AR as an effective solution generation approach that augments the search space of existing methods.

\section{Related Work}
\label{sec:related_work}

\textbf{Autonomous science}: A current open area of research is exploring whether AI can help augment or improve scientific research. Some works utilize end-to-end approaches that start with an initial objective, dataset, or codebase and allow LLM agents to iterate on ideas, execute code, and draft scientific reports \citep{sakana-ai-scientist, sakana-ai-scientist-v2, kosmos, denario}. Other works explore more specific parts of the scientific pipeline, like hypothesis \citep{autodiscovery, towards-an-ai-coscientist} or idea generation \citep{can-llms-generate-novel-research-ideas, towards-execution-grounded, research-agent, llms-for-automated, virsci}. Given the complexity of biomedical research, many autonomous science systems tackle problems in this field, like COVID-19 nanobody design or drug repurposing \citep{virtual-lab, virtual-biotech, cellvoyager, towards-an-ai-coscientist}. However, the success of all these frameworks relies heavily on their ability to consistently generate high-quality and diverse candidate solutions to explore.

\textbf{Diversity mode collapse}: A significant barrier to open-ended LLM generation is diversity mode collapse, which is a well-documented phenomenon where outputs converge to a narrow set of high-probability generations \citep{artificial-hivemind, hypospace, digital-red-queen, can-llms-explore-in-context, homogenization-effects, shared-imagination, price-of-format, strong-model-collapse, forcing-diffuse-distributions, base-models-beat-aligned-models}. In open-ended research tasks, this collapse is particularly problematic as it effectively limits the search space of potential discoveries. Prior work has attempted to mitigate mode collapse through various prompting, decoding, and training techniques \citep{verbalized-sampling, quality-diversity, min-p, diverse-beam-search, diverse-preference-optimization, dive}. However, approaches that enforce structural and semantic diversity remain an open opportunity.

\textbf{Analogical reasoning}: Previous research from the field of cognitive science has defined the paradigm of ``analogical reasoning'' (AR), in which knowledge is translated across different domains through shared relational structure \citep{analogical-reasoning}. Various works have demonstrated that LLMs exhibit a latent ability to perform AR \citep{emergent-analogical-reasoning, llms-as-models-for-analogical-reasoning, llms-as-analogical-reasoners, relevant-or-random, ar-for-strategic-decisions}. While LLM-driven AR has been successfully applied in various tasks like software engineering \citep{u2f}, mathematical proof generation \citep{ar-for-proof-generation}, and automated ICD-10 coding \citep{ar-for-icd10-coding}, its application to solution generation in autonomous science is unexplored.

\section{Methods}

\subsection{Open-ended Solution Generation}

We define the task of open-ended solution generation. Let $\mathcal{P}$ denote the space of research problems and $\mathcal{S}$ denote the space of solutions. Given a research problem $p \in \mathcal{P}$, generate a solution $s \in \mathcal{S}$ for $p$. The solution $s$ is evaluated on various axes (see Section \ref{sec:experiments}) as a possible approach to solve $p$. This task is more constrained than other autonomous science tasks like hypothesis generation, which is an intentional choice in order to isolate the benefits of analogical reasoning. We outline our formulation of analogical reasoning in the following sections.

\subsection{Definition of Analogy}

We utilize the definition of analogy established by the structure-mapping framework from cognitive science \citep{structure-mapping}. Structure-mapping theory defines analogies as ``implicit rules for mapping knowledge about a base domain into a target domain.'' This knowledge is mapped through relational structure shared between both the base domain (from which the knowledge originates) to the target domain (where the knowledge is applied to). Within the structure-mapping framework, objects are defined as entities that function as wholes at a given level of organization. Relations are predicates that take in two or more arguments to express how objects interact. Attributes are predicates that take in a single argument to describe an object's properties. While shared relations map knowledge across domains, object attributes are not mapped. See example analogy in Figure \ref{fig:overview}A.

Given a research problem $p$, we define the set of objects $O_p$, which consists of the objects that are utilized within the research problem. We define the set of relations $R_p$, which consists of the relationships between the objects within $O_p$. In our work, we represent analogies as two components:

\begin{enumerate}
    \item \textbf{Object Mappings} between the domains
    \item \textbf{Shared Relations} between the domains
\end{enumerate}

\subsection{Analogical Reasoning}
\label{sec:analogical_reasoning}

We utilize this definition of analogy in our formulation of analogical reasoning. For our task of open-ended solution generation, we define analogical reasoning as the ability to generate analogies from a research problem $p$ to cross-domain research problems that exhibit similar relational structure. We implement AR as a two-step process consisting of ``extraction'' and ``search'' steps. The extraction step extracts analogical components from $p$ and generates analogies to other domains. The search step uses these analogies to search for novel solutions to $p$.

\textbf{Extraction}: Given a research problem $p$, we extract the set of objects $O_p$ and the set of relations $R_p$ from $p$, and generate an analogy $a \in \mathcal{A}_{\mathrm{cross}}$. The set $\mathcal{A}_{\mathrm{cross}} \subset \mathcal{A}$ denotes the space of analogies to cross-domain research problems, with $\mathcal{A}$ representing the universal space of all possible analogies (including in-domain analogies). The generated analogy $a$ contains a set of objects $O_a$ to which the original set of problem objects $O_p$ map. In addition, the generated analogy $a$ includes a set of relations $R_a$ such that $R_a \subseteq R_p$. This allows for partial relational mapping, which means the analogy is not required to have complete coverage of the original problem's relations. Each analogy $a$ consists of two components:

\begin{enumerate}
    \item \textbf{Object Mappings} ($M: O_p \leftrightarrow O_a$) between problem domain and analogous domain
    \item \textbf{Shared Relations} ($R_a$) between problem domain and analogous domain
\end{enumerate}

So, an equivalent formulation for the analogy is $a = (M, R_a)$, or the object mappings and shared relations. For each research problem $p$, this extraction step is performed by one call to an extraction agent (LLM) with temperature=$1.0$.

\begin{table*}[t]
\centering
\footnotesize
\renewcommand{\arraystretch}{0.85} 
\linespread{0.85}\selectfont       
\begin{tabular}{p{12.5cm}}
\toprule
\textbf{Analogy 1:} drug delivery $\Leftrightarrow$ environmental engineering \\
\hspace{0.5em}Object Mappings: \newline \begin{tabular}{@{\hspace{1.5em}}l}$\bullet$ implantable delivery system $\leftrightarrow$ leaking underground storage tank \\ $\bullet$ anticancer drug $\leftrightarrow$ contaminant chemical \\ $\bullet$ tumor microenvironment $\leftrightarrow$ subsurface soil layers \\ $\bullet$ healthy tissue $\leftrightarrow$ protected water resources\end{tabular} \\
\hspace{0.5em}Shared Relations: \newline \begin{tabular}{@{\hspace{1.5em}}l}$\bullet$ a stationary source releases substance into heterogeneous medium \\ $\bullet$ substance disperses forming concentration gradient \\ $\bullet$ transport pathways remove substance from area\end{tabular} \\
\textbf{Candidate Solution:} MODFLOW with MT3DMS Reactive Transport Modeling for Multi-Species Contaminant Plume Prediction from Injection Wells \citep{mt3dms} \\
\hspace{1.5em}$\bullet$ \begin{minipage}[t]{11.5cm}MT3DMS (Modular Three-Dimensional Multispecies Transport Model) coupled with MODFLOW simulates advection, dispersion, and sorption of reactive contaminants from point sources in heterogeneous aquifer systems. The model uses finite-difference methods to solve 3D advection-dispersion equations with reaction terms, incorporating spatial heterogeneity in hydraulic conductivity, porosity, and sorption coefficients to predict concentration distributions over time.\end{minipage} \\
\midrule
\textbf{Analogy 2:} biomedical signal processing $\Leftrightarrow$ seismology \\
\hspace{0.5em}Object Mappings: \newline \begin{tabular}{@{\hspace{1.5em}}l}$\bullet$ neural sources $\leftrightarrow$ earthquake hypocenters \\ $\bullet$ scalp electrodes $\leftrightarrow$ seismometers \\ $\bullet$ brain tissue $\leftrightarrow$ earth crust and mantle \\ $\bullet$ electrical signals $\leftrightarrow$ seismic waves\end{tabular} \\
\hspace{0.5em}Shared Relations: \newline \begin{tabular}{@{\hspace{1.5em}}l}$\bullet$ internal sources generate propagating waves within a volume conductor \\ $\bullet$ waves travel through heterogeneous medium with attenuation \\ $\bullet$ external sensors measure superimposed signals\end{tabular} \\
\textbf{Candidate Solution:} Double-Difference Earthquake Location Algorithm \citep{dd, hypodd} \\
\hspace{1.5em}$\bullet$ \begin{minipage}[t]{11.5cm}The double-difference algorithm minimizes residuals between observed and theoretical travel-time differences for pairs of earthquakes at common stations, rather than absolute travel times. This approach cancels out errors from unknown velocity structure along common ray paths, significantly improving relative location accuracy for clustered seismic events. The method iteratively refines hypocenter locations by solving a large sparse linear system using LSQR or conjugate gradient methods.\end{minipage} \\
\bottomrule
\end{tabular}
\caption{Example analogies generated by AR that reveal structural similarity between disparate domains through object mappings and shared relations. These analogies are leveraged to transfer novel candidate solutions from an analogous domain back to the problem domain (e.g. applying a transport model from environmental engineering to drug delivery, as shown in Analogy 1).}
\label{tab:analogy_examples}
\end{table*}

\textbf{Search}: Next, we utilize the extracted analogies to find candidate solutions for the research problem $p$. For each analogy $a$, we provide a search agent (LLM) with $p$ and $a$, and prompt it to utilize $a$ to search for solutions in other domains. This search step is performed by one call to the search agent with temperature=$1.0$. In our work, one solution is generated per analogy. The output consists of the generated solution $s \in \mathcal{S}$. See Section \ref{sec:ar_prompts} for the prompts used in the extraction and search steps.

\section{Evaluation Setup}
\label{sec:experiments}

Within the task of open-ended solution generation, we evaluate analogical reasoning (AR) on 3 axes (generation diversity, solution novelty, analogy quality) to assess the utility of the output generations.

\subsection{Evaluation Axes}
\label{sec:evaluation_axes}
The generation diversity axis evaluates the semantic diversity of the candidate domains and solutions discovered by different settings. We use the Vendi Score \citep{vendi} as the primary metric of evaluating semantic diversity. The Vendi Score is a diversity metric that takes as input any similarity function and can be interpreted as the effective number of unique elements in a set. We use cosine similarity as our similarity function. A higher Vendi Score signifies more diverse generations. The solution novelty axis evaluates the novelty of the candidate solutions. We utilize an LLM-judged novelty score that is assessed using relevant literature. We show that the LLM-judge novelty scorer aligns with human-annotated novelty scores (see Section \ref{sec:novelty_scoring_human_annotations}). In addition to the LLM-judged novelty scores, we perform a human study to determine which generated solutions are perceived as more novel by human evaluators (see Section \ref{sec:solution_novelty_human_annotations}). The analogy quality axis evaluates the analogies generated by the different settings. We define high-quality analogies as original and structurally sound mappings that successfully transfer knowledge across disparate domains. To quantify this, we evaluate analogy quality using three LLM-judged metrics: Structural Depth (measures insightfulness of object mappings), Domain Distance (measures distance between problem domain and analogous domain), and Analogy Novelty (measures originality of analogy). We show the LLM-judged analogy quality scores align with human-annotated preferences (see Section \ref{sec:analogy_quality_human_annotations}). See Section \ref{sec:evaluation_setup_details} for details.

\subsection{Baselines}
\label{sec:baselines}
We evaluate three settings for solution generation: AR, a cross-domain baseline, and a no-domain baseline. As mentioned in Section \ref{sec:analogical_reasoning}, AR generates analogies to different domains and uses those analogies to find solutions to the research problem. The cross-domain baseline also finds solutions from different domains to the research problem, but does not explicitly generate analogies to do so. The no-domain baseline simply finds solutions to the research problem without any domain constraints. We select these baselines because they represent natural and strong approaches that reflect how users typically prompt LLMs for generating solutions to open-ended problems. Note that AR uses 2 LLM calls per generation (1 call for the extraction step, 1 call for the search step). The cross-domain baseline also uses 2 LLM calls per generation (1 call to generate candidate domains, 1 call for the search step). The no-domain baseline uses 1 LLM call (1 call for the search step). See Section \ref{sec:ar_prompts} and Section \ref{sec:baseline_prompts} for prompts used for the 3 settings. For all 3 settings, each candidate generation consists of a solution and the domain that the solution originates from (see Section \ref{sec:example_generations} for example generations). Across all settings, we evaluate 3 LLMs: Claude Sonnet 4.5, GPT-5.2, and Gemini 3 Flash. All LLMs use temperature=1.0.

\section{Quantitative Results}
\label{sec:quantitative_results}

\subsection{AR mitigates LLM mode collapse}
\begin{figure}[h]
\begin{center}
\includegraphics[scale=0.607]{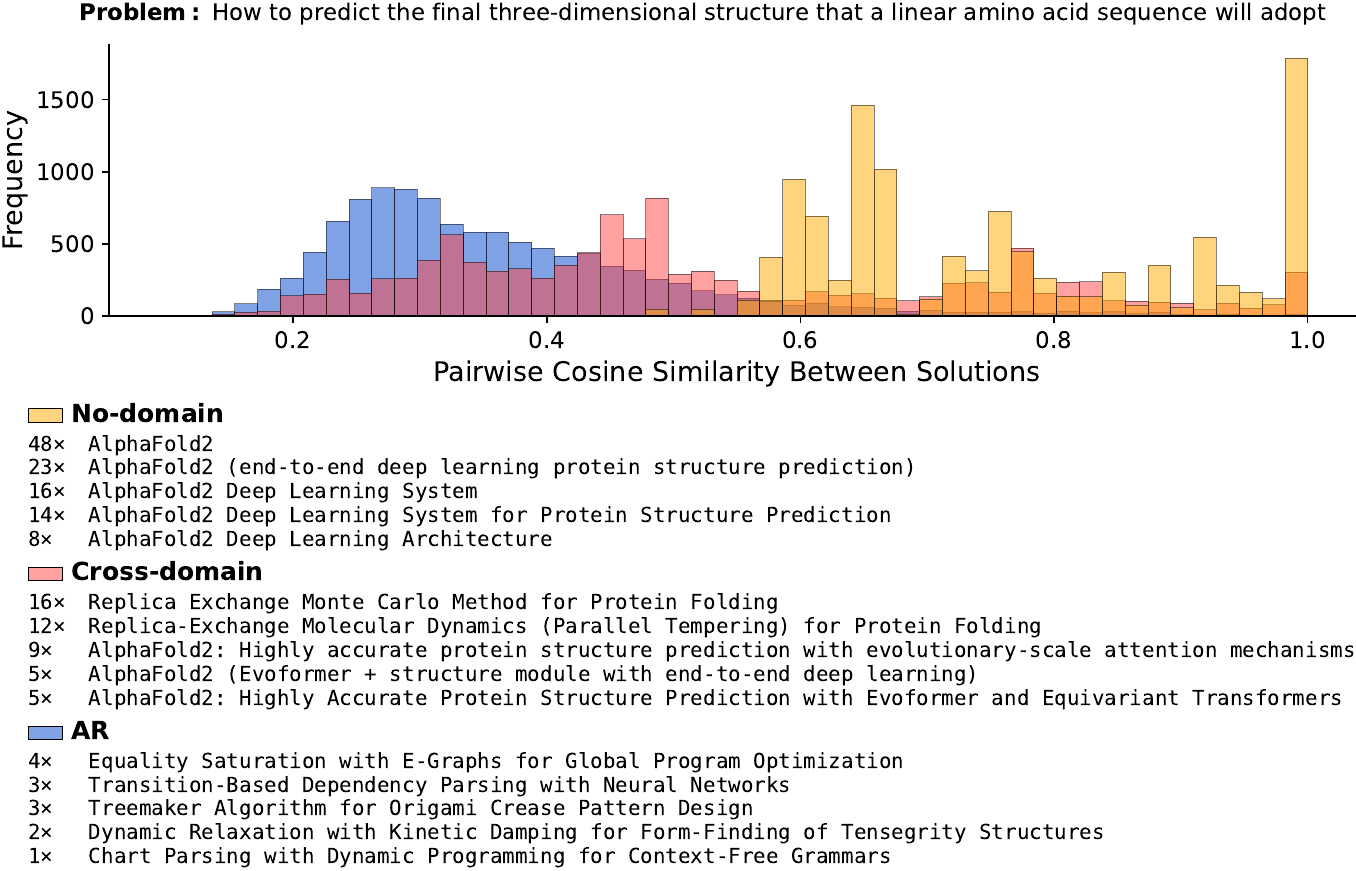}
\caption{Example solutions for one research problem. (top) Distribution of pairwise cosine similarities between 150 solutions aggregated across all LLMs within each of the three settings (no-domain, cross-domain, AR). Lower similarities indicate more semantically diverse solutions. (bottom) Top-5 most frequent solutions and their counts across all three settings. Baseline methods capture the most obvious solutions (e.g. AlphaFold2), but AR discovers a broader set of novel solutions that augments the search space of existing approaches.} 
\label{fig:mode_collapse}
\end{center}
\end{figure}

When evaluating generation diversity across the 3 settings, we find that both baseline settings demonstrate significant mode collapse in the domain and solution space. In the aggregated evaluation where there are 150 generations per problem, the no-domain and cross-domain baselines achieve average domain Vendi Scores of 2.64 and 2.84 (see Figure \ref{fig:overview}C) and solution Vendi Scores of 5.81 and 8.34 (see Figure \ref{fig:overview}C, Table \ref{table:diversity}). In addition, only approximately 5\% of the generated domains (7-8 unique domains) and approximately 57-66\% of the generated solutions (86-98 unique solutions) were unique. The significant mode collapse of the baseline settings is evident in the high cosine similarity between solutions (see Figure \ref{fig:mode_collapse}). 

This finding is concerning, as it demonstrates that LLMs in their current form lack the capacity to propose a wide variety of approaches to research problems. The no-domain baseline predictably collapses strongly, as it was simply tasked with finding solutions to the research problem. Some research problems in the AR Dataset have canonical solutions, like AlphaFold for the protein folding problem, which the search agent would likely collapse to when prompted for solutions to the problem. Rather surprisingly, the cross-domain baseline, which is tasked to search for solutions in other domains, still collapses to low semantic diversity generations.

We find that AR helps mitigate this mode collapse. AR significantly outperforms the baselines by 100-115\% on domain Vendi Score (see Figure \ref{fig:domain_vendi_bar_chart}) and 90-173\% on solution Vendi Score (see Figure \ref{fig:overview}C, Table \ref{table:diversity}). In addition, it finds nearly 3.5x more unique domains and 1.3x more unique solutions than the next best baseline. This improvement in generation diversity is likely due to the search space of analogies being larger than the search space of solution recall (which the no-domain baseline does) or cross-domain solution recall (which the cross-domain baseline does).

\subsection{AR finds more novel solutions}
Though we show that AR generates more diverse solutions than the baseline settings, we also want to evaluate whether AR-generated solutions are better on other axes. We choose solution novelty as one metric to assess on. For all 3 LLMs, we find AR solutions are scored as consistently more novel than the baseline solutions (see Table \ref{table:novelty}). On the stratified novelty score, AR outperforms the baseline methods on Claude (scores of 1.28 $<$ 3.12 $<$ 5.54, with no-domain $<$ cross-domain $<$ AR), GPT (1.98 $<$ 3.69 $<$ 6.14), and Gemini (2.11 $<$ 4.68 $<$ 6.43). On the binary novelty score, we also see the same trends of AR finding the most novel solutions, followed by cross-domain, and lastly no-domain. This trend is present for all 3 LLMs (Claude: 1.6\% $<$ 21.6\% $<$ 50.4\%, GPT: 9.0\% $<$ 28.2\% $<$ 58.8\%, Gemini: 8.9\% $<$ 37.8\% $<$ 68.9\%). The binary novelty scores show that over half of the solutions that AR finds are novel while as little as a quarter of the cross-domain baseline solutions found are novel. The no-domain baseline rarely finds novel solutions. Across the 3 LLMs, we find that Gemini generates the most novel solutions, followed by GPT, and then Claude. 

In addition to using LLM-judged novelty scores, we evaluate solution novelty using human annotations. We conduct a human pairwise preference study in which humans determine whether a randomly selected AR solution or a cross-domain solution is more novel. We choose the cross-domain baseline to compare against as it is the strongest baseline. We find that humans score AR-generated solutions as more novel 78\% of the time with a moderate inter-annotator agreement rate ($\kappa=0.445$). AR solutions are also perceived as practical, as they are scored as a reasonable approach to the problem 67\% of the time. We envision AR as an alternative solution generation approach that expands the search space to uncover more novel solutions. See Section \ref{sec:solution_novelty_human_annotations} for further details.

\begin{table}[h]
\centering
\begin{tabular}{l c c c c c c}
\toprule
& \multicolumn{2}{c}{No-domain} & \multicolumn{2}{c}{Cross-domain} & \multicolumn{2}{c}{AR} \\
\cmidrule(lr){2-3} \cmidrule(lr){4-5} \cmidrule(lr){6-7}
LLM & \makecell[c]{Stratified \\ ($\uparrow$)} & \makecell[c]{Binary \\ ($\uparrow$)} & \makecell[c]{Stratified \\ ($\uparrow$)} & \makecell[c]{Binary \\ ($\uparrow$)} & \makecell[c]{Stratified \\ ($\uparrow$)} & \makecell[c]{Binary \\ ($\uparrow$)} \\
\midrule
Claude & $1.28$ & $1.6\%$ & $3.12$ & $21.6\%$ & $\textbf{5.54}$ & $\textbf{50.4\%}$ \\
GPT & $1.98$ & $8.8\%$ & $3.69$ & $27.2\%$ & $\textbf{6.14}$ & $\textbf{58.8\%}$ \\
Gemini & $2.11$ & $8.6\%$ & $4.68$ & $37.9\%$ & $\textbf{6.43}$ & $\textbf{68.8\%}$ \\
\bottomrule
\end{tabular}
\caption{Average of mean per-problem novelty scores. Evaluated across 50 research problems with 5 solutions generated per problem. Results are shown for three settings (no-domain, cross-domain, AR) for three LLMs (Claude, GPT, Gemini) and assessed across two LLM-judge scoring prompts (stratified and binary).}
\label{table:novelty}
\end{table}

\subsection{AR analogies are high-quality}
We also evaluate the quality of the analogies used to generate the solutions. On the Structural Depth metric, all settings perform comparably with scores ranging from 7.06 to 8.60 across the 3 different LLMs (see Figure \ref{fig:analogy_results_all}, Table \ref{table:analogy}). However, when comparing the Domain Distance between the problem domain and analogous domain, we see that AR discovers analogies from domains that are consistently further than the other settings (with no-domain $<$ cross-domain $<$ ground-truth $<$ AR for all 3 LLMs) with Domain Distance scores (6.99, 6.83, 7.53 for Claude, GPT, Gemini) over 3x higher than the no-domain baseline (2.29, 3.08, 3.37 for Claude, GPT, Gemini). AR also finds more novel analogies than all other settings, even the ground-truth analogy from the AR Dataset. AR analogies score the highest on novelty (4.82, 5.27, 5.58) across all 3 LLMs. See Section \ref{sec:additional_plots} for full results. In addition, the LLM-judged analogy quality scores align strongly with human-annotated preferences, with human-LLM agreement rates as high as 88.6\% for the Domain Distance metric and 75.7\% for the Novelty metric. See Section \ref{sec:analogy_quality_human_annotations} for further details.

\section{Case Studies}

We have shown that AR-generated solutions are more diverse and novel than the baseline LLM-generated solutions. However, we also want to demonstrate the practical feasibility and efficacy of the candidate solutions. We evaluate AR across four biomedical problems by generating and implementing one promising solution per problem. These solutions are straightforward to implement using a coding agent (Claude Code using Opus 4.6) with minimal human guidance. The performance of each implemented approach is measured against published baselines. See Section \ref{sec:additional_case_study_details} for further implementation details and results.

\begin{figure}[h]
\begin{center}
\includegraphics[scale=0.403]{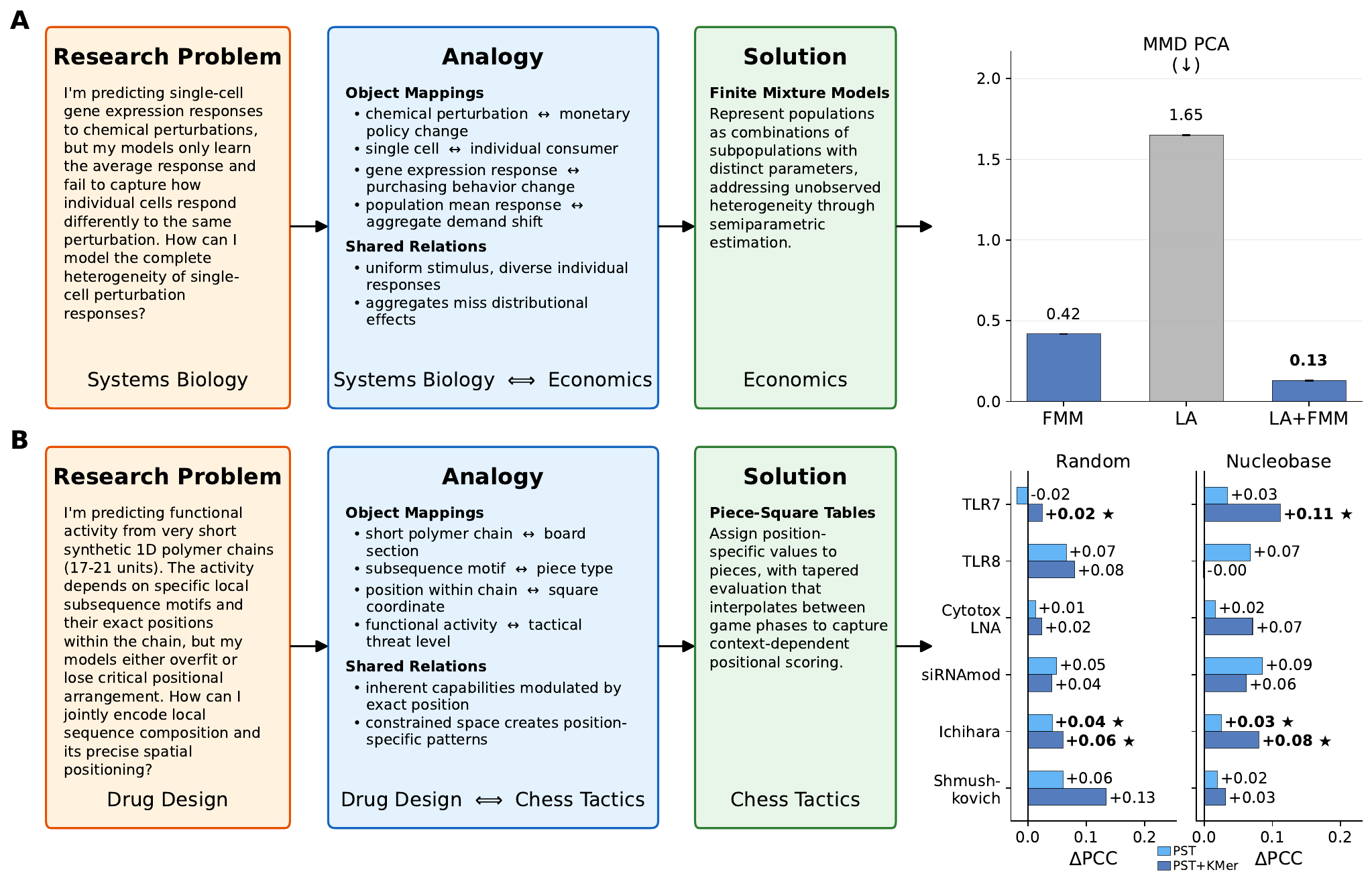}
\caption{Analogical reasoning pipeline for case study \#1 and case study \#4. (A) Perturbation effect prediction: (left) AR generates an analogy between systems biology and economics. (right) FMM and LA+FMM compared to the LA baseline on the MMD PCA distributional metric. (B) Oligonucleotide property prediction: (left) AR generates an analogy between drug design and chess. (right) $\Delta$PCC (Pearson correlation coefficient) gain over the Linear baseline for PST and PST+Kmer across 6 datasets and 2 splits, star denotes new SOTA.} 
\label{fig:case_study_results}
\end{center}
\end{figure}

\subsection{Case Study \#1: Perturbation Effect Prediction}

We choose the problem of perturbation effect prediction for our first case study. This problem is currently an open area of research, as models frequently fail to beat simple baselines \citep{DL-based-predictions-dont-outperform}. The PerturBench benchmark compiles 6 perturbational datasets and performs comprehensive benchmarking to assess model performance \citep{perturbench}. For the purposes of our proof-of-concept case study, we evaluate the performance of our AR-generated solution on the Srivatsan20 dataset, which is 1 of the 6 datasets provided in the benchmark. The Srivatsan20 dataset evaluates on the task of covariate transfer, in which models are trained on a set of observed cell lines and tested on held-out cell lines. The paper notes that existing models tend to learn the average expression response to perturbations and fail to capture the complete heterogeneity of effects. We task AR to find solutions that allow for modeling the full distribution of the perturbational effect.

To solve the problem, AR generated an analogy that mapped the systems biology domain, where the source problem of perturbation effect prediction resides, to the economics domain. Similar to how cell lines respond differently to perturbations, consumers also respond differently to economic policy changes. In both settings, aggregate measures can miss the individual heterogeneity of behavior. Using this analogy, AR proposed applying the economics approach of finite mixture models (FMM) to perturbation effect prediction. This finite mixture model approach achieves state-of-the-art performance on the MMD PCA distributional distance metric with a value of 0.42 (see Figure \ref{fig:case_study_results}A; lower MMD is better). In addition, when combining the finite mixture model approach with the Latent Additive (LA) baseline from the paper, we further improve performance over the vanilla LA baseline on MMD PCA from 1.65 to 0.13 (a nearly 13x improvement) while maintaining comparable performance on prediction accuracy metrics  (Cosine LogFC, Cosine LogFC rank, RMSE mean, RMSE mean rank).

\subsection{Case Study \#2: Cell-Cell Communication Inference}
For our second case study, we explore the task of cell-cell communication inference from single-cell RNA-seq data. The OpenProblems cell-cell communication ligand-target benchmark evaluates whether methods can predict the presence of ligand-receptor interactions \citep{openproblems}. Existing approaches evaluated in the benchmark predominantly determine interactions based on ligand-receptor co-expression. However, co-expression only indicates whether the signaling molecules are present, not whether the interaction activates a downstream pathway. We tasked AR with discovering novel approaches that predict interactions by considering the downstream transcriptional response.

AR generated an analogy from the telecommunications domain that equates a radio receiver quantifying a structured signal against a background noise floor to a true ligand-receptor interaction producing a transcriptional response that rises above background expression. Using this analogy, AR discovered signal-to-noise ratio (SNR) analysis, which determines true signals by accepting only those that exceed a certain confidence threshold. We evaluated the SNR solution against all 14 baselines from the OpenProblems benchmark on the AUPRC and odds ratio metrics. The AR-proposed SNR approach achieves state-of-the-art performance on AUPRC with a value of 0.248, while remaining competitive on odds ratio (see Figure \ref{fig:cs_ccc_figure}).

\subsection{Case Study \#3: Brain Region Interaction}
For our third case study, we apply AR to brain region interaction data. The data includes simultaneous neural activity readings from the left and right hemispheres of the brain \citep{modularity}. A key task in the original paper explores isolating cross-hemisphere coupling signals to understand interhemispheric interactions during mouse decision-making. However, extracting this signal is difficult due to the strong autocorrelation inherent in neural data. To address this, the approach in the paper predicts temporal changes in activity rather than raw values and employs a two-stage Ridge regression model. For this case study, we tasked AR with discovering novel approaches to account for this temporal autocorrelation while accurately isolating the cross-hemisphere coupling.

AR generated an analogy from communication networks that draws comparisons between traffic flow in a network and the signal propagating throughout the brain. Using this analogy, AR discovered the Peter and Clark Momentary Conditional Independence (PCMCI) method which uses causal discovery techniques to account for autocorrelation. When applied to a set of 18 neural activity sessions, the PCMCI approach successfully isolates the coupling signal, which is demonstrated by statistically significant and positive correlation values for all 18 sessions. We compare the isolated signal to the signal derived from the paper's approach, and both methods showed high agreement with a Spearman correlation of $\rho=0.729$. The PCMCI coupling signal also achieves comparable predictive power to the Ridge baseline for downstream behavioral outcomes like selectivity recovery ($\rho=0.335$ for PCMCI vs. $\rho=0.317$ from the original paper, see Figure \ref{fig:cs3_figure}).

\subsection{Case Study \#4: Oligonucleotide Property Prediction}
For our last case study, we explore the task of oligonucleotide property prediction. The OligoGym benchmark contains 12 datasets of oligonucleotide sequences paired with therapeutic efficacy and safety measurements \citep{oligogym}. The authors benchmark a variety of classical and deep learning methods on this task of sequence property prediction. One of the key issues noted in the paper is that larger deep learning models overfit on the short oligonucleotide sequences or other models are unable to model positional information. We explore whether AR can suggest novel approaches to solve this issue.

 AR discovered a very interesting analogy from chess tactics, in which the value of a chess piece varies based on its position on the board. In the oligonucleotide sequence space, therapeutic properties are similarly determined by the position of certain motifs in the sequence. Using this analogy, AR discovered the chess solution of piece-squared tables (PST) with tapered evaluation, which interpolates value tables for certain pieces between opening and endgame phases based on the game phase. The PST implementation for sequence property prediction parallels this by interpolating between a certain position's proximity to the terminus or center of the sequence. In doing so, the model encodes a structural prior to allow more flexibility to capture position-dependent effects.

We evaluate on the 6 smallest datasets (with $<$1000 measurements) using a simple linear regression model. This allows a fair comparison to the Linear model from the benchmark and isolates the predictive power of the new PST features. Using a simple model ensures any performance gains are from the improved features rather than a complex model's capacity. We assess performance on both the Random and Nucleobase splits. We find the PST solution achieves consistent improvements over the Linear model while using the same architecture, improving on 5 of 6 datasets in the Random setting and 6 of 6 datasets in the Nucleobase setting. When we further augment the approach to include both the PST features and the k-mer features used in the paper, we achieve SOTA performance on 2 of 6 datasets in both the Random and Nucleobase settings (see Figure \ref{fig:case_study_results}B).

\section{Discussion}
\label{sec:discussion}

In this work, we introduce analogical reasoning (AR) as a structured approach to open-ended solution generation. By mapping shared relational structure across disparate domains, candidate generations from AR are significantly more diverse, novel, and high-quality than baselines. AR represents a vital step toward the larger goal of bridging isolated domains to drive scientific innovation. As the field of autonomous science accelerates, frameworks will require highly novel and diverse hypotheses to avoid convergence to canonical solutions. AR can serve as a ``diversity engine'' for these future systems that augments the search space of existing solution generation methods. In addition, the advent of high-quality code executors shows promise for high-speed implementation and evaluation of candidate solutions. AR is also cheap to run at approximately \$0.02 per problem, making it practical to deploy at scale.

\textbf{Limitations}: The work contains several limitations. First, while AR vastly expands the search space for novel ideas, not every proposed analogy translates to a feasible solution. Evaluating the real-world viability of these hypotheses remains a bottleneck. In addition, the current AR pipeline isolates the solution generation step, but future work should integrate AR into an end-to-end execution-grounded pipeline. Lastly, while our evaluation and curated AR Dataset are currently scoped to biomedical problems, the AR framework is domain-agnostic. Expanding AR to other scientific disciplines is a promising area of future work.

\newpage
\section*{Acknowledgments}
We thank Owen Queen for helpful feedback on the work and Samuel Alber, Peter Eckmann, Arushi Gupta, Sophia Kivelson, Zijian Carl Ma, Alan Mao, Jiacheng Miao, and Jake Silberg for serving as human annotators. A.S. is supported by NSF GRFP grant DGE-2146755.

\newpage
\bibliography{neurips_2026}
\bibliographystyle{neurips_2026}

\newpage
\appendix

\section{Reproducibility Statement}
\label{sec:reproducibility_statement}
We describe the evaluation setup for our analyses and methods in the main text and Appendix. All LLM prompts used are provided in Section \ref{sec:prompts}. In addition, we include the codebase for our work and the AR Dataset at this link: https://github.com/andrew7shen/ar\_science

\section{Evaluation Setup Details}
\label{sec:evaluation_setup_details}

We outline additional implementation details for the 3 evaluation axes of generation diversity, solution novelty, and analogy quality.

\subsection{Generation Diversity Details}
\label{sec:generation_diversity_details}
We use cosine similarity as the kernel function when calculating the Vendi Score for generation diversity. We also report the raw number of unique domains and solutions, which is determined by counting unique generations after converting to lowercase and stripping whitespace. From the 266 papers in the AR Dataset (see Section \ref{sec:ar_dataset_details}), we sample 50 papers to create the evaluation set. We constrain the size of our evaluation set to 50 papers to limit API costs required for the analysis.

For each of the 50 papers, we utilize the research problem from the paper and feed the research problem as input into the solution generation settings (no-domain, cross-domain, AR). We generate 50 candidate generations for each of the 50 papers, which results in 2500 candidate generations per setting. For each candidate generation, we separately embed the domain and the solution using the OpenAI embedding model ``text-embedding-3-small''. We calculate the Vendi Score across all domain embeddings (domain Vendi Score) and solution embeddings (solution Vendi Score) for all candidate generations for each setting. We calculate the average domain Vendi Score and solution Vendi Score across all 50 papers. This analysis is repeated with each of the 3 LLMs (Claude, GPT, Gemini). To evaluate semantic diversity across LLMs, we also compute Vendi Scores across an aggregated set of generations across all 3 LLMs.

\subsection{Solution Novelty Details}
\label{sec:solution_novelty_details}
The setup for the solution novelty evaluation is similar to that of the generation diversity evaluation in Section \ref{sec:generation_diversity_details}, as we use the same set of sampled 50 papers from the 266 total papers in the AR Dataset as our evaluation set. For the solution novelty axis, we generate 5 candidate solutions for each of the 50 papers, which results in 250 candidate solutions per setting. To assess the novelty of each of the candidate solutions, we determine if the candidate solution has ever been applied to the specific research problem, which we refer to as the ``solution transfer''.

\textbf{Retrieval and Re-ranking}: We utilize a retrieval and re-ranking strategy, inspired by the Idea Novelty Checker \citep{idea-novelty-checker}, in which we first retrieve a large set of relevant papers and then re-rank those papers based on embedding similarity to the proposed solution transfer. For the retrieval step, we query the Semantic Scholar API to pull relevant papers. We task an LLM (Claude Haiku 4.5, temperature=0.0) to generate 3 Semantic Scholar queries for each solution transfer. The aim is to generate queries that will find papers similar to the proposed solution transfer. See Section \ref{sec:query_generation_prompt} for the query generation prompt used. Each query is executed separately on the Semantic Scholar API and returns the top-15 most relevant papers. After the retrieval step, we perform embedding re-ranking of all the queried papers (a maximum of 45 papers with top-15 from each of the 3 queries) by calculating the cosine similarity between the embeddings of the solution transfer and each of the queried papers. We utilize the SPECTER \citep{specter} embeddings of each of the queried papers. To generate the embedding of the solution transfer, we task an LLM (Claude Haiku 4.5, temperature=0.0) to rewrite the solution transfer into a paper-like title and abstract and embed that with SPECTER to enable fair comparison. See Section \ref{sec:rewritten_solution_transfer_prompt} for the rewritten solution transfer prompt used. We return the final top-10 papers after the re-ranking step.

\textbf{Novelty Scoring}: We use the set of top-10 relevant papers to assess the novelty of the proposed solution transfer. The top-10 relevant papers and the solution transfer are passed to an LLM-judge (Claude Sonnet 4.5, temperature=0.0) to assess the novelty of the solution transfer. We utilize 2 different LLM judge prompts to assess the novelty scores, one using a stratified scoring range (from 1-10) and one using a binary scale (True/False). We show that both LLM-judge prompts align with human-annotated novelty scores (see Section \ref{sec:novelty_scoring_human_annotations}). See Section \ref{sec:novelty_judge_prompts} for the LLM-judge novelty prompts used.

\subsection{Analogy Quality Details}
\label{sec:analogy_quality_details}
For the analogy quality evaluation, we utilize LLM-judged scores. To generate the LLM-judge scores, we feed in the research problem, problem domain, analogous domain, and the analogy used into an LLM-judge (Claude Sonnet 4.5, temperature=0.0). We show that the analogy quality LLM-judge prompt aligns with human-annotated preferences (see Section \ref{sec:analogy_quality_human_annotations}). See Section \ref{sec:analogy_judge_prompts} for the LLM-judge analogy quality prompt used. However, though each AR solution contains an analogy, each cross-domain and no-domain solution may not necessarily contain an analogy. We include an additional analogy extraction step for the cross-domain and no-domain baselines, in which the analogy is extracted if the candidate solution includes analogical components (see Section \ref{sec:analogy_extraction} for more details). We evaluate the analogies from 5 candidate solutions generated for each of the 50 papers. We also include the ground-truth analogy used in each AR Dataset paper as an additional baseline to compare to the AR- and baseline-generated analogies (see Section \ref{sec:analogy_extraction} for more details).

\section{Human Annotations}
\label{sec:human_annotations}
\subsection{Solution novelty human annotations}
\label{sec:solution_novelty_human_annotations}

In addition to LLM-judged novelty scores, we evaluate solution novelty using human annotations. We conduct a human pairwise preference study in which human annotators determine which of the two provided solutions is more novel. For a given research problem, one solution generated by AR is compared against one solution generated by the cross-domain baseline. Within each pair, the presentation order of the two solutions is randomized to control for position bias. To generate the solution pairs, we sampled one random AR solution and one random cross-domain baseline solution for each of the 50 research problems in our evaluation set, resulting in 50 pairs of solutions. We recruited four biomedical AI researchers as annotators. The four annotators are split into two groups of two, with each group evaluating 25 pairs. For each pair, annotators select which solution is more novel, and also whether each of the proposed solutions seems like a reasonable approach to the problem. See ``Solution Preference Study Instructions'' below for the instructions we provide to annotators.

For the human annotation analysis, we calculate novelty rate, reasonable rate, and Cohen's $\kappa$. Novelty rate measures the fraction of human votes that scored the AR solution as more novel than the cross-domain baseline solution. Reasonable rate measures the fraction of human votes that scored the given solution (AR or cross-domain baseline) as reasonable. Cohen's $\kappa$ measures inter-annotator agreement corrected for chance and is averaged across annotator pairs.

We find that the AR solutions are scored as more novel than the cross-domain baseline solutions 78\% of the time. In addition, the human annotators exhibit moderate inter-annotator agreement on which solutions are more novel ($\kappa$=0.445). AR solutions are scored as reasonable 67\% of the time while cross-domain baseline solutions are scored as reasonable 86\% of the time. The inter-annotator agreement for reasonable rate is also much lower for AR than cross-domain baseline solutions ($\kappa$ of 0.155 and 0.416 respectively). This demonstrates that the increased perceived novelty of the AR solutions may come at a cost of being a less reasonable solution (see Table \ref{table:solution_human_eval}). We envision AR not as a replacement for baseline search approaches, but as an alternative approach that expands the search space to uncover novel and diverse solutions.

\begin{table}[h]
\centering
\setlength{\tabcolsep}{4pt}
\renewcommand{\arraystretch}{1.3}
\begin{tabular}{l c c}
\toprule
~ & \makecell[c]{{Rate} \\ ($\uparrow$)} & \makecell[c]{{Cohen's $\kappa$} \\ ($\uparrow$)} \\
\midrule
Novelty (AR pref.) & 0.78 & 0.445 \\
Reasonable (AR) & 0.67 & 0.155 \\
Reasonable (Cross-domain) & 0.86 & 0.416 \\
\bottomrule
\end{tabular}
\caption{Results from human pairwise preference study on solution novelty. We report novelty rate, reasonable rate, and Cohen's $\kappa$.}
\label{table:solution_human_eval}
\end{table}

\begin{promptbox}[title={Solution Preference Study Instructions}]
\begin{lstlisting}[
    basicstyle=\ttfamily\scriptsize,
    breaklines=true,
    columns=fullflexible,
    keepspaces=true,
    frame=none,       % Turn off listing frame (box handles it)
    aboveskip=0pt,    % Remove extra gap at top
    belowskip=0pt     % Remove extra gap at bottom
]
You are evaluating solutions proposed to solve scientific problems.

This is a pairwise comparison task. For each question, you will see a scientific problem and two proposed solutions (Solution A and Solution B).
  
For each pair, you will answer 3 questions:
  
Q1: Which solution is more novel? (Solution A / Solution B)
- A novel solution proposes an original, non-obvious, or creative approach.
- Consider: Would a typical researcher think of this? Does it draw from an unexpected domain or technique?
       
Q2: Does Solution A seem like a reasonable approach? (Yes / No)
Q3: Does Solution B seem like a reasonable approach? (Yes / No)
- A reasonable solution is one that makes sense as an approach to the problem.                                    
- An unreasonable solution would be nonsensical, irrelevant, or clearly inapplicable to the problem.                
- Note: Solutions may come from other scientific domains. A solution is reasonable if the underlying idea is relevant, even if it would need modification to apply directly.
  
What you will see for each solution:
- Title: Name of the proposed solution
- Source Domain: The field the solution originates from
- Description: A brief explanation of the solution
    
NOTE: These solutions span diverse scientific domains. You are not expected to be an expert in every field. Please use your best judgment.
\end{lstlisting}
\end{promptbox}

\subsection{Solution novelty scorer evaluation}
\label{sec:novelty_scoring_human_annotations}

To validate our LLM-judge novelty scorer, we compare the LLM-judged scores to the human-annotated novelty scores from the AI Researcher dataset \citep{can-llms-generate-novel-research-ideas}. The dataset includes NLP research ideas annotated by expert NLP researchers. These annotations include novelty scores from 1-10, with 10 denoting highly novel ideas. We utilize the 49 AI-generated and 49 human-written ideas (N=98 total ideas) from the dataset for our evaluation. We assess our stratified and binary LLM-judge novelty scoring methods by passing each of the 98 NLP research ideas through our novelty scoring pipeline to generate an LLM-judged novelty score. For the stratified scorer, we measure the Spearman and Pearson correlation between LLM-judge and human novelty scores. Since each idea has multiple annotations, we report results using four aggregation strategies for the human ground truth: median, mean, minimum, and maximum. For the binary scorer, we binarize the human ground truth (scores >5 as novel) and report classification metrics across each aggregation strategy. We also compute a human-human baseline by measuring pairwise agreement between reviewers on the same idea (see Table \ref{table:novelty_correlation}). We note that the LLM-judge to human agreement is computed against the aggregated human score, while the human-human baseline uses individual pairwise comparisons. This may slightly favor the LLM-judge comparison. We also measure the mean absolute difference (MAD) between the LLM-judge and individual human annotators (see Table \ref{table:novelty_mad}) to assess the degree of disagreement between scores. Lower MAD indicates better agreement.

We find that the stratified scorer achieves moderate Spearman correlation ($\rho$=0.36-0.43) with human annotations for the median, mean, and minimum aggregation strategies. This correlation value is comparable to correlation values from the Agentic Reviewer (human-human $\rho$=0.41, LLM-human $\rho$=0.42) work that measures similar LLM-judge to human agreement \citep{agentic-reviewer}. The binary scorer is comparable to the human-human baseline on accuracy and F1 score for the median and mean aggregation strategies, which indicates that the LLM-judge scores align with human experts.  The MAD for the stratified scorer (1.70 for N=91) is comparable to that of the human baseline (1.73 for N=91), which demonstrates that the LLM-judge scores disagree with individual human experts no more than human experts disagree with each other.

\begin{table}[h]
\centering
\setlength{\tabcolsep}{4pt}
\renewcommand{\arraystretch}{1.3}
\begin{tabular}{l c c c c}
\toprule
~ & \multicolumn{2}{c}{{Stratified}} & \multicolumn{2}{c}{{Binary}} \\
\cmidrule(lr){2-3} \cmidrule(lr){4-5}
{Aggregation} & \makecell[c]{{Spearman $\rho$} \\ ($\uparrow$)} & \makecell[c]{{Pearson $r$} \\ ($\uparrow$)} & \makecell[c]{{Acc.} \\ ($\uparrow$)} & \makecell[c]{{F1} \\ ($\uparrow$)} \\
\midrule
Median & 0.358$^{*}$ & 0.359$^{*}$ & 0.602 & 0.672 \\
Mean   & 0.411$^{*}$ & 0.386$^{*}$ & 0.633 & 0.695 \\
Min    & 0.427$^{*}$ & 0.422$^{*}$ & 0.469 & 0.435 \\
Max    & 0.245$^{*}$ & 0.250$^{*}$ & 0.612 & 0.725 \\
\midrule
Human & --- & --- & 0.571 & 0.562 \\
\bottomrule
\end{tabular}
\caption{LLM-judge novelty scorer agreement with human experts for the stratified and binary scoring methods across four aggregation strategies (median, mean, min, max). For the stratified scorer, we report correlation metrics (Spearman, Pearson). For the binary scorer, we report classification metrics (accuracy, F1) and a human-human pairwise baseline. $^{*}p < 0.05$.}
\label{table:novelty_correlation}
\end{table}

\begin{table}[h]
\centering
\setlength{\tabcolsep}{4pt}
\renewcommand{\arraystretch}{1.3}
\begin{tabular}{l c c c c}
\toprule
~ & \multicolumn{2}{c}{\makecell[c]{{All} \\ {(N=98)}}} & \multicolumn{2}{c}{\makecell[c]{{$\geq$2 Reviews} \\ {(N=91)}}} \\
\cmidrule(lr){2-3} \cmidrule(lr){4-5}
{Method} & \makecell[c]{{MAD} \\ ($\downarrow$)} & \makecell[c]{{SD} \\ ($\downarrow$)} & \makecell[c]{{MAD} \\ ($\downarrow$)} & \makecell[c]{{SD} \\ ($\downarrow$)} \\
\midrule
Stratified         & 1.73 & 1.14 & 1.70 & 1.11 \\
Binary & 4.59 & 3.69 & 4.40 & 3.54 \\
\midrule
Human     & ---  & ---  & 1.73 & 1.19 \\
\bottomrule
\end{tabular}
\caption{Mean absolute difference (MAD) and standard deviation (SD) between LLM-judge (stratified and binary methods) novelty scores and individual human reviewer scores. We report results for all ideas (N=98) and for ideas with 2 or more annotators (N=91).}
\label{table:novelty_mad}
\end{table}

\subsection{Analogy quality scorer evaluation}
\label{sec:analogy_quality_human_annotations}

To validate our LLM-judged analogy quality metrics, we conduct a human pairwise preference study. We evaluate our three primary LLM-judged metrics (Structural Depth, Domain Distance, Novelty). For each of the three metrics, we construct 20 high-vs-low pairs by selecting analogies from the top and bottom quartiles of the LLM score distribution for that metric, resulting in 60 total pairs. Within each pair, the presentation order of the two analogies is randomized to control for position bias. We recruited four biomedical AI researchers as annotators. The four annotators are split into two groups of two, with each group evaluating 30 pairs (10 per metric). For each pair, annotators select which analogy is better on the given metric, or indicate that the analogies are ``about equal''. We provide annotators with metric definitions and calibration examples before the evaluation (see ``Analogy Preference Study Instructions'' below for the instructions we provide to annotators).

For the human annotation analysis, we calculate accuracy, Cohen's $\kappa$, and equal rate. Accuracy measures the fraction of non-equal human votes that match the LLM ranking direction. Cohen's $\kappa$ measures inter-annotator agreement corrected for chance and is averaged across annotator pairs. Equal rate measures the fraction of votes where annotators chose ``about equal''.

We find that Domain Distance and Novelty achieve high human-LLM agreement (accuracy of 0.886 and 0.757 respectively) with moderate inter-annotator agreement ($\kappa$ of 0.451 and 0.327 respectively). Structural Depth shows weaker agreement (accuracy=0.600) with low inter-annotator agreement ($\kappa$=-0.133) which suggests this metric is more difficult to assess by humans (see Table \ref{table:analogy_human_eval}). All metrics show low equal rate values, which demonstrates that the annotators were able to discriminate high-vs-low analogies.

\begin{table}[h]
\centering
\setlength{\tabcolsep}{4pt}
\renewcommand{\arraystretch}{1.3}
\begin{tabular}{l c c c}
\toprule
~ & \makecell[c]{{Accuracy} \\ ($\uparrow$)} & \makecell[c]{{Cohen's $\kappa$} \\ ($\uparrow$)} & \makecell[c]{{Equal Rate} \\ ($\downarrow$)} \\
\midrule
Structural Depth & 0.600 & -0.133 & 0.125 \\
Domain Distance & 0.886 & 0.451 & 0.125 \\
Novelty & 0.757 & 0.327 & 0.075 \\
\bottomrule
\end{tabular}
\caption{LLM-judged analogy quality metric agreement with human pairwise preference annotations across three metrics (Structural Depth, Domain Distance, Novelty).  We report accuracy, Cohen's $\kappa$, and equal rate.}
\label{table:analogy_human_eval}
\end{table}

\begin{promptbox}[title={Analogy Preference Study Instructions}]
\begin{lstlisting}[
    basicstyle=\ttfamily\scriptsize,
    breaklines=true,
    columns=fullflexible,
    keepspaces=true,
    frame=none,       % Turn off listing frame (box handles it)
    aboveskip=0pt,    % Remove extra gap at top
    belowskip=0pt     % Remove extra gap at bottom
]
You are evaluating analogies proposed to solve scientific problems.

This is a pairwise preference ranking task. For each question, you will see two analogies (Analogy A and Analogy B) side by side, and you will be asked to evaluate them on one specific metric.

You have 3 choices for each pair:
  - Analogy A is better on the given metric
  - Analogy B is better on the given metric
  - About equal

Each pair is evaluated on only one metric (not all three). The metric being evaluated is shown at the top of each pair.

The 3 metrics:

STRUCTURAL DEPTH: How well-defined and meaningful are the object mappings?
  0 = Vague or superficial mappings with little explanatory power
  10 = Exceptional mappings with deep insight into structural correspondence

DOMAIN DISTANCE: How far apart are the problem domain and analogous domain?
  0 = Same or closely related domains
  10 = Highly disparate domains with no obvious overlap

NOVELTY: How original and insightful is this analogy?
  0 = Very common or obvious analogy
  10 = Groundbreaking insight, highly innovative

What you will see for each analogy:
  - Problem: the scientific problem being addressed
  - Domain Transfer: Problem Domain -> Analogous Domain
  - Object Mappings: how concepts in the problem domain map to concepts in the analogous domain
  - Shared Relations: the structural relationships that hold across both domains

NOTE: While these analogies span diverse scientific domains, please evaluate them based on your scientific judgment. You are not expected to be an expert in every field represented. In addition, each analogy contains detailed information (problem, mappings, rationales, shared relations). You do not need to read every detail in depth to make your judgment, as a quick scan is often sufficient to get a sense of the analogy's quality on the given metric.
\end{lstlisting}
\end{promptbox}

\section{Additional Case Study Details}
\label{sec:additional_case_study_details}
We provide additional details on the four case studies, including the research problem used, the AR-generated analogies, and the solution. To generate solutions, we use the same AR pipeline used for the evaluation tasks in Section \ref{sec:quantitative_results}, but replace the search agent (Claude, GPT, or Gemini) with Perplexity Deep Research. We make this choice because the case studies require implementing solutions from the cited sources and codebases, and Perplexity Deep Research's grounded search reduces the source and codebase hallucinations we observed with the other LLMs. To implement the solution, we provide a coding agent (Claude Code using Opus 4.6) with the solution description, cited sources, and cited codebases. The coding agent is tasked to implement the solution with little human guidance.

\subsection{Case Study \#1: Perturbation Effect Prediction Details}

\textbf{Research Problem}: I'm predicting single-cell gene expression responses to chemical perturbations, but my models only learn the average response and fail to capture how individual cells respond differently to the same perturbation. I need to predict the full diversity of cell-to-cell responses, not just the population mean. How can I model the complete heterogeneity of single-cell perturbation responses?

\textbf{Analogy}: AR generated an analogy from economics. Similar to how consumers with heterogeneous economic profiles respond differently to the same monetary policy change, individual cells with distinct internal states respond differently to the same chemical perturbation. In both settings, aggregate measures (population mean response or aggregate demand) obscure the distributional effects that are critical for understanding the system's behavior.

\textbf{Solution}: Using this analogy, AR discovered finite mixture models (FMMs) from econometrics, which represent populations as combinations of distinct subpopulations. We implement the FMM by fitting a Gaussian mixture model to the perturbation data and sample predictions with cell-type-specific mixing proportions. We compare the standalone FMM, the reproduced LA baseline from PerturBench, and a hybrid LA+FMM that recenters the GMM components onto the LA mean prediction to combine accurate mean estimation with distributional modeling. See Table \ref{table:cs1_results} for full results.

\begin{table}[h]
\centering
\setlength{\tabcolsep}{4pt}
\renewcommand{\arraystretch}{1.3}
\resizebox{\textwidth}{!}{%
\begin{tabular}{l  c c c c c}
\toprule
{Method} & \makecell[c]{Cosine LogFC \\ ($\uparrow$)} & \makecell[c]{Cosine LogFC rank \\ ($\downarrow$)} & \makecell[c]{MMD PCA \\ ($\downarrow$)} & \makecell[c]{RMSE mean \\ ($\downarrow$)} & \makecell[c]{RMSE mean rank \\ ($\downarrow$)} \\
\midrule
FMM & 0.08 $\pm$ 0.000 & 0.36 $\pm$ 0.000 & 0.42 $\pm$ 0.000 & 0.028 $\pm$ 0.000 & 0.32 $\pm$ 0.000 \\
LA & \textbf{0.47 $\pm$ 0.013} & \textbf{0.16 $\pm$ 0.013} & 1.65 $\pm$ 0.002 & \textbf{0.017 $\pm$ 0.000} & \textbf{0.16 $\pm$ 0.010} \\
LA+FMM & 0.45 $\pm$ 0.008 & 0.17 $\pm$ 0.012 & \textbf{0.13 $\pm$ 0.002} & 0.018 $\pm$ 0.000 & \textbf{0.16 $\pm$ 0.010} \\
\bottomrule
\end{tabular}}
\caption{Perturbation effect prediction performance on 5 metrics (Cosine LogFC, Cosine LogFC rank, MMD PCA, RMSE mean, RMSE mean rank) from the PerturBench benchmark on the Srivatsan20 dataset. FMM is the AR-proposed finite mixture model solution, LA is the latent additive baseline from the PerturBench paper, and LA+FMM combines both.}
\label{table:cs1_results}
\end{table}

\subsection{Case Study \#2: Cell-Cell Communication Inference}

\textbf{Research Problem}: I'm predicting which ligand-receptor interactions between cell types are functionally active using single-cell RNA-seq data. Methods that score interactions by ligand-receptor co-expression capture whether the molecules are present, but not whether the downstream signaling pathway is actually activated in the target cells. How can I predict which interactions drive downstream transcriptional responses, not just which molecules are co-expressed?

\textbf{Analogy}: AR generated an analogy from telecommunications. In the telecommunications domain, a radio receiver quantifies the structured radio signal against a background noise floor. Transmissions that exceed a threshold signal-to-noise ratio are accepted. Similarly, true ligand-receptor interactions in noisy single-cell RNA-seq data must produce a response that rises above the background expression.

\begin{figure}[h]
\begin{center}
\includegraphics[scale=0.465]{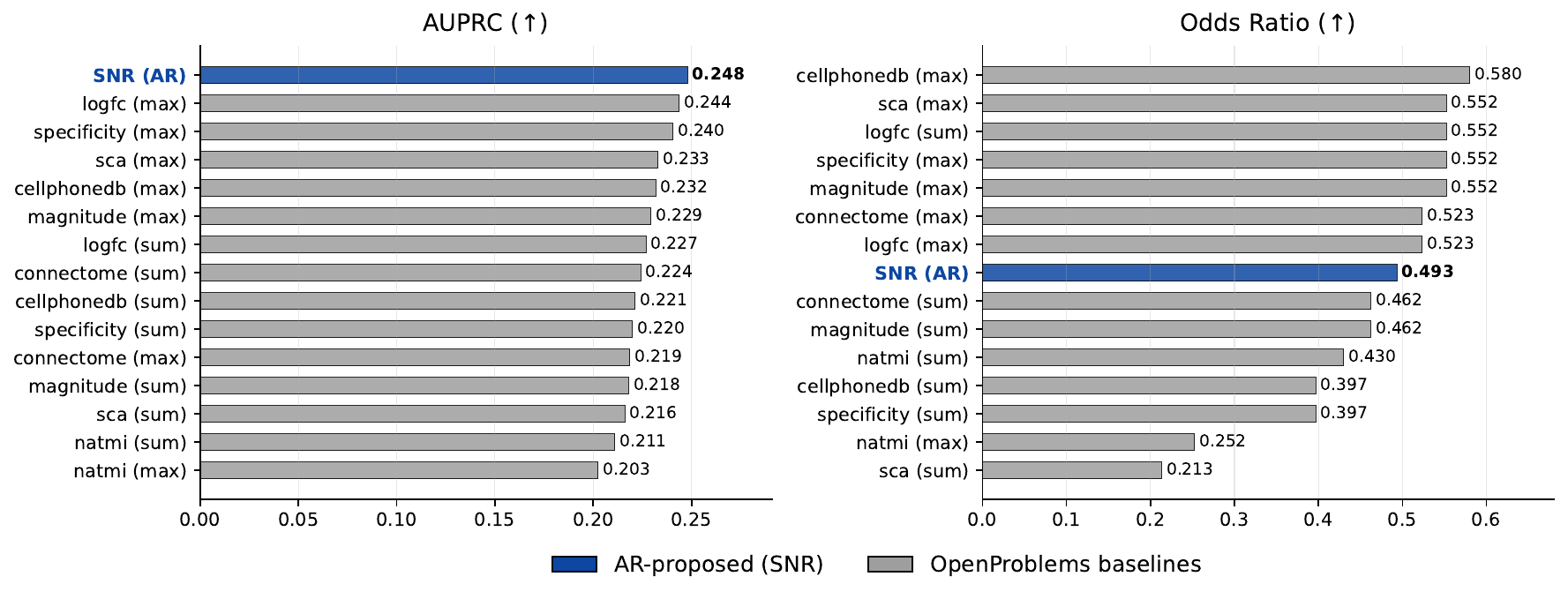}
\caption{Cell-cell communication inference results on 2 metrics (AUPRC, odds ratio) from the OpenProblems ligand-target benchmark. Performance is shown for the AR-proposed signal-to-noise ratio (SNR) solution against baselines.} 
\label{fig:cs_ccc_figure}
\end{center}
\end{figure}

\textbf{Solution}: Using this analogy, AR discovered signal-to-noise ratio (SNR) analysis, which quantifies a structured signal against a background noise floor. We treat ligand-specific downstream transcription factor (TF) activity in the target cell type as the structured signal and the population-level distribution of the same TF activity as the noise floor. We calculate the z-score against a permutation null as the confidence measure. Ligand-to-TF mappings are derived by performing breadth-first search over KEGG signal-transduction pathways, and per-cell TF activity is computed using CollecTRI regulon data. We return the geometric mean of the co-expression and transcription factor activity signal. We evaluate on the OpenProblems cell-cell communication ligand-target benchmark against all 14 baselines. See Figure \ref{fig:cs_ccc_figure} for full results.

\subsection{Case Study \#3: Brain Region Interaction Details}

\textbf{Research Problem}: I have simultaneous recordings from the left and right hemispheres of the same brain region. Each hemisphere's activity is highly autocorrelated, so its own past dominates any predictive model and masks the coupling signal from the other hemisphere. How can I isolate the influence one hemisphere has on the other when each hemisphere's own autocorrelation dominates the signal?

\textbf{Analogy}: AR generated an analogy from communication networks. Detecting inter-network routing influence amid local traffic patterns is similar to detecting cross-hemisphere coupling amid neural autocorrelation. In both settings, two systems with strong autocorrelation (local traffic patterns or neural activity) have weak mutual influences (inter-network routing effects or cross-hemisphere coupling) that are masked by the dominant self-predictive signal.

\textbf{Solution}: Using this analogy, AR discovered PCMCI (Peter-Clark Momentary Conditional Independence), a constraint-based causal discovery method. PCMCI handles autocorrelation through partial correlation conditioning. We implement PCMCI with partial correlation tests directly on the raw neural activity projections (8 variables: CD + 3 orthogonal PCs per hemisphere) which outputs a causal graph. We report the PCMCI cross-hemisphere coupling strength as the mean absolute partial correlation across all significant cross-hemisphere edges in the graph, minus the mean from a trial-shuffled null distribution. We also reproduce the paper's two-stage Ridge regression baseline, which predicts temporal changes after removing within-hemisphere prediction components. 

\begin{figure}[h]
\begin{center}
\includegraphics[scale=0.3]{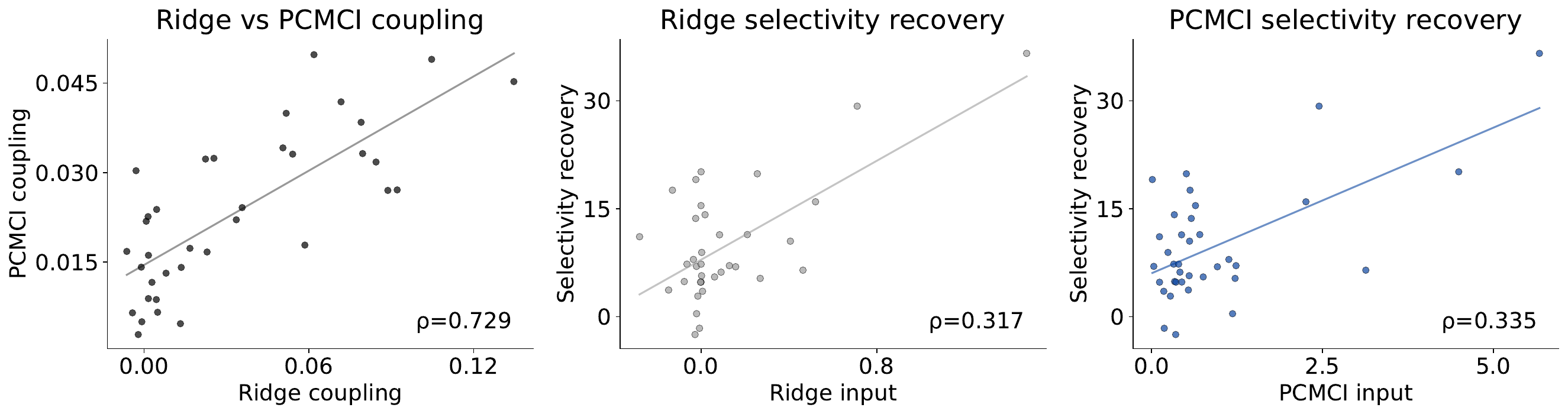}
\caption{Brain region interaction results across N=36 hemisphere-sessions. (left) Spearman correlation between strength of Ridge and PCMCI cross-hemisphere coupling signals. (middle, right) Spearman correlation between Ridge and PCMCI coupling signal and selectivity prediction.} 
\label{fig:cs3_figure}
\end{center}
\end{figure}

In addition, we evaluate how well each approach's measure of inferred interhemispheric input predicts the mouse's decision after a targeted brain perturbation. When comparing the predictive power of both approaches, the baseline Ridge approach yielded a Spearman correlation of $\rho=0.317$ (95\% bootstrap CI $[-0.018,0.612]$) while the PCMCI method yielded $\rho=0.335$ (95\% CI $[-0.025,0.637]$, N=36 hemisphere-sessions, 10,000 paired resamples). These results demonstrate that the PCMCI method discovered by AR achieves comparable predictive performance to the Ridge baseline (see Figure \ref{fig:cs3_figure}).

\subsection{Case Study \#4: Oligonucleotide Property Prediction Details}

\textbf{Research Problem}: I'm predicting functional activity from very short synthetic 1D polymer chains (17-21 units). The activity depends on specific local subsequence motifs and their exact positions within the chain, but my models either overfit or lose critical positional arrangement. How can I jointly encode local sequence composition and its precise spatial positioning?

\textbf{Analogy}: AR generated an analogy from chess tactics. The value of a chess piece on its exact square position on the board, similar to how the functional contribution of a nucleotide motif depends on its exact position in the sequence. In both settings, identical elements (pieces or motifs) have position-dependent functional effects within a constrained spatial arrangement.

\textbf{Solution}: Using this analogy, AR discovered piece-square tables (PST) with tapered evaluation from chess. PSTs assign specific values to each piece based on its exact board square, and tapered evaluation interpolates between two sets of tables (opening and endgame phases) based on game state. We implement PST features through separate terminal and central one-hot encoding vectors by weighting features proportionally to their distance from the sequence center. We evaluate with a linear regression model to match the OligoGym Linear baseline, and also test a PST+KMer variant that concatenates PST features with the benchmark's k-mer count features. See Table \ref{table:cs4_results} for full results.

\begin{table}[h]
\centering
\setlength{\tabcolsep}{4pt}
\renewcommand{\arraystretch}{1.3}
\resizebox{\textwidth}{!}{%
\begin{tabular}{l l c c c c}
\toprule
{Split} & {Dataset} & {Linear} & {PST} & {PST+KMer} & {Benchmark SOTA} \\
\midrule
 & TLR7 & 0.77 $\pm$ 0.02 & 0.75 $\pm$ 0.04 & \textbf{0.79 $\pm$ 0.04} & 0.78 $\pm$ 0.04 \\
 & TLR8 & 0.54 $\pm$ 0.13 & 0.61 $\pm$ 0.03 & 0.62 $\pm$ 0.09 & \textbf{0.68 $\pm$ 0.08} \\
Random & Cytotox LNA & 0.79 $\pm$ 0.02 & 0.80 $\pm$ 0.02 & 0.81 $\pm$ 0.02 & \textbf{0.91 $\pm$ 0.02} \\
 & siRNAmod & 0.62 $\pm$ 0.06 & 0.67 $\pm$ 0.03 & 0.66 $\pm$ 0.03 & \textbf{0.69 $\pm$ 0.03} \\
 & Ichihara & 0.54 $\pm$ 0.07 & 0.58 $\pm$ 0.04 & \textbf{0.60 $\pm$ 0.07} & 0.56 $\pm$ 0.02 \\
 & Shmushkovich & 0.24 $\pm$ 0.09 & 0.30 $\pm$ 0.04 & 0.37 $\pm$ 0.05 & \textbf{0.50 $\pm$ 0.08} \\
\midrule
 & TLR7 & 0.68 $\pm$ 0.07 & 0.71 $\pm$ 0.07 & \textbf{0.79 $\pm$ 0.05} & 0.73 $\pm$ 0.10 \\
 & TLR8 & 0.59 $\pm$ 0.08 & 0.66 $\pm$ 0.11 & 0.59 $\pm$ 0.08 & \textbf{0.70 $\pm$ 0.06} \\
Nucleobase & Cytotox LNA & 0.64 $\pm$ 0.16 & 0.66 $\pm$ 0.05 & 0.71 $\pm$ 0.06 & \textbf{0.72 $\pm$ 0.11} \\
 & siRNAmod & 0.44 $\pm$ 0.11 & 0.53 $\pm$ 0.07 & 0.50 $\pm$ 0.04 & \textbf{0.53 $\pm$ 0.09} \\
 & Ichihara & 0.49 $\pm$ 0.09 & 0.52 $\pm$ 0.03 & \textbf{0.57 $\pm$ 0.11} & \textbf{0.57 $\pm$ 0.12} \\
 & Shmushkovich & 0.20 $\pm$ 0.08 & 0.22 $\pm$ 0.08 & 0.23 $\pm$ 0.03 & \textbf{0.42 $\pm$ 0.14} \\
\bottomrule
\end{tabular}}
\caption{Oligonucleotide property prediction PCC (Pearson correlation coefficient) on 6 datasets from the OligoGym benchmark for Random and Nucleobase splits. PST and PST+KMer are the AR-proposed piece-square table inspired models, Linear is the linear baseline from the paper, and Benchmark SOTA is the best reported PCC across all OligoGym models.}
\label{table:cs4_results}
\end{table}

\section{Analogy Extraction}
\label{sec:analogy_extraction}
Each AR solution contains an explicitly structured analogy with object mappings and shared relations, as these components are produced as part of the AR pipeline. However, the cross-domain and no-domain baseline solutions are generated without explicit analogical structure. To fairly evaluate analogy quality across all settings, we include an additional analogy extraction step for baseline solutions.

For each baseline solution, we prompt an LLM (Claude Sonnet 4.5, temperature=0.0) to extract the cross-domain analogy (represented by object mappings and shared relations) given the research problem, problem domain, analogous domain, and the solution (see Section \ref{sec:analogy_extraction_prompts} for the prompt used).

A baseline solution is considered to contain a valid analogy if the extraction step produces a non-empty set of object mappings. Solutions where the analogous domain (from which the solution originates) matches the problem domain (where the solution is applied) return empty object mappings and are excluded from analogy scoring.

The ground-truth analogy from the AR Dataset paper is also extracted in the same format. We prompt an LLM (Claude Sonnet 4.5, temperature=0.0) to extract the object mappings and shared relations to allow for fair comparison against the analogies from other settings (see Section \ref{sec:analogy_extraction_prompts} for the prompt used).

\section{AR Dataset Details}
\label{sec:ar_dataset_details}
\subsection{AR Dataset}
\label{sec:ar_dataset}
We compile a dataset of research papers that resulted in cross-domain breakthroughs by using analogical reasoning. To curate the Analogical Reasoning Dataset, or ``AR Dataset'', we utilize an LLM to discover published papers that fit the criteria of utilizing analogical reasoning. We use Perplexity Sonar Pro with web and literature search tools as the LLM for AR Dataset curation due to its strength in literature search with citations. The LLM was tasked to discover up to 15 papers where an analogy from a base domain was the key insight to enable a solution in the target domain. This discovery step was performed for every pairwise combination of base and target domains, which were drawn from a master set of 48 non-biomedical base domains and 47 biomedical target domains (see Section \ref{sec:domains} for list of domains). The base and target domains sets were chosen to cover a wide range of research domains, but are not exhaustive.

After discovery of a set of candidate papers, the source of each paper was verified using the Semantic Scholar \citep{semantic-scholar} and arXiv literature search APIs. Candidate papers were rejected if they were unable to be verified by the literature search APIs. For each successfully verified paper, we use an LLM (Claude Sonnet 4.5) to extract relevant analogical metadata (including research problem, problem domain, analogous domain, analogy used, etc.) and assess the difficulty of each analogy for each paper. We also preprocess the research problem for each of the 50 evaluation papers before passing it to the solution generation pipeline. Since the original research problems are extracted diretly from the published paper, they frequently leak information about the analogous domain into the problem. Feeding such problem statements directly into the search agents would exacerbate mode collapse in the baseline methods, which would not be fair when comparing to AR. We use an LLM (Claude Sonnet 4.5) to rewrite each original problem to not include any hints from the analogous domain. The rewritten problem is the input provided to all three settings (no-domain, cross-domain, AR). We use the same rewritten problems across all 3 LLMs evaluated. See Section \ref{sec:ar_dataset_prompts} for the prompts used. We currently constrain the AR Dataset to solely biomedical problems to limit the scope of the dataset.

The final AR Dataset contains 266 curated examples of analogical reasoning in biomedicine. We use the research problems from the samples in the AR Dataset as a source of problems to evaluate solution diversity and novelty (Section \ref{sec:generation_diversity_details}, Section \ref{sec:solution_novelty_details}). We also use the analogies used in the AR Dataset samples in our analogy evaluation (Section \ref{sec:analogy_quality_details}). We release the AR Dataset as a resource to the community.

\subsection{Master set of domains}
\label{sec:domains}
To generate candidate papers for the AR Dataset, we search for papers between 48 non-biomedical domains and 47 biomedical domains (see Table \ref{table:domains}).

\begin{table}[htbp]
\centering
\scriptsize
\begin{tabular}{p{3.0cm} p{3.3cm} | p{3.0cm} p{2.9cm}}
\multicolumn{2}{c|}{\normalsize\textbf{Non-biomedical Domains}} & \multicolumn{2}{c}{\normalsize\textbf{Biomedical Domains}} \\[0.5em] 
statistical physics & quantum mechanics & immunology & neuroscience \\
optics and photonics & fluid dynamics & cancer biology & pharmacology \\
topology & optimization theory & systems biology & epidemiology \\
dynamical systems & game theory & medical imaging & infectious disease \\
information theory & computational complexity & cardiology & metabolic disease \\
formal verification & automata theory & regenerative medicine & developmental biology \\
cryptography & distributed systems & endocrinology & genetics and genomics \\
control systems & signal processing & protein engineering & drug delivery systems \\
robotics & computer vision & structural biology & synthetic biology \\
network theory & machine learning architectures & bioinformatics & tissue engineering \\
error correction codes & database systems & cell signaling & biomechanics \\
compiler optimization & scheduling algorithms & toxicology & molecular diagnostics \\
telecommunications & power systems & hematology & biochemistry \\
structural engineering & behavioral economics & genetic disorders & biomedical sensors \\
network sociology & organizational theory & virology & stem cell biology \\
hydrology & geomorphology & microbiome & drug discovery \\
seismology & economics & vascular biology & nephrology \\
materials science & linguistics & ophthalmology & pathology \\
urban planning & architecture & anesthesiology & radiology \\
music theory & archaeology & nanomedicine & epigenetics \\
ecology & population dynamics & aging/gerontology & wound healing \\
thermodynamics & nonlinear dynamics & organ transplantation & autoimmune disease \\
stochastic processes & swarm intelligence & psychiatric disorders & pain medicine \\
agent-based modeling & collective behavior & reproductive biology & \\
\end{tabular}
\caption{List of the 48 non-biomedical domains and 47 biomedical domains used to find candidate papers for the AR Dataset.}
\label{table:domains}
\end{table}

\section{Prompts}
\label{sec:prompts}
We include the prompts used by the LLMs in our work. The AR, cross-domain, and no-domain prompts include instructions to perform each method of solution search. All settings also include the same instructions to find valid sources and codebases for each solution.

\subsection{AR Prompts}
\label{sec:ar_prompts}

\begin{promptbox}[title={AR Extraction Prompt}]
\begin{lstlisting}[
    basicstyle=\ttfamily\scriptsize,
    breaklines=true,
    columns=fullflexible,
    keepspaces=true,
    frame=none,       % Turn off listing frame (box handles it)
    aboveskip=0pt,    % Remove extra gap at top
    belowskip=0pt     % Remove extra gap at bottom
]
You are analyzing a biomedical research problem to find analogous solutions in other domains through explicit object-to-object mappings.

**Problem:**
{problem_text}

**Your task:**
Extract the problem's key objects and relations, then generate analogies to {num_domains} other domains with explicit object mappings.

**Return a JSON object with:**

1. "problem_summary": A concise 1-2 sentence summary
2. "problem_objects": Array of key objects/entities with their functional roles
3. "problem_relations": Array of core relational structures between objects
4. "analogies": Array of {num_domains} analogies, each with:
   - "target_domain": The domain name (e.g., "computer_science", "logistics")
   - "analogy_title": A descriptive title for this analogy
   - "object_mappings": Array of source-to-target mappings with rationale
   - "shared_relations": The relational structure preserved across domains
5. "key_terms": Array of {min_key_terms}-{max_key_terms} important terms/concepts
6. "target_domains": Array of the {num_domains} domain names (from analogies)

**Map objects by FUNCTION, not surface similarity.** "Delivers payload" is a good mapping basis; "is liquid" is not.

**Example format:**
```json
{{
  "problem_summary": "Brief description of the biomedical problem",
  "problem_objects": [
    {{"name": "object_A", "role": "functional role of A"}},
    {{"name": "object_B", "role": "functional role of B"}}
  ],
  "problem_relations": [
    "key relation between objects"
  ],
  "analogies": [
    {{
      "target_domain": "domain_name",
      "analogy_title": "Title describing the analogy",
      "object_mappings": [
        {{"source": "object_A", "target": "analogous_A", "mapping_rationale": "why these map functionally"}},
        {{"source": "object_B", "target": "analogous_B", "mapping_rationale": "why these map functionally"}}
      ],
      "shared_relations": "the relational structure preserved in this domain"
    }}
  ],
  "key_terms": ["term1", "term2", "term3"],
  "target_domains": ["domain_name"]
}}
```

Return ONLY the JSON object, no other text.

\end{lstlisting}
\end{promptbox}

\begin{promptbox}[title={AR Search Prompt}]
\begin{lstlisting}[
    basicstyle=\ttfamily\scriptsize,
    breaklines=true,
    columns=fullflexible,
    keepspaces=true,
    frame=none,       % Turn off listing frame (box handles it)
    aboveskip=0pt,    % Remove extra gap at top
    belowskip=0pt     % Remove extra gap at bottom
]
You are finding solutions from {domain} that could be applied to a biomedical problem through analogical reasoning.

**Biomedical Problem:**
{problem_summary}

**Analogy to {domain}:**
{analogy_title}

**Object Mappings (biomedical -> {domain}):**
{object_mappings}

**Shared Relational Structure:**
{shared_relations}

**Key Concepts:**
{key_terms}

**IMPORTANT:** Use the object mappings above to guide your search. Find solutions in {domain} that work with the TARGET objects (right side of mappings) and preserve the shared relational structure. Do NOT search for direct biomedical solutions.

**OUTPUT CONCISENESS:**
- Keep descriptions to 2-3 sentences, focus on core algorithm/method only
- Avoid verbose background - provide only technical essentials for analogical reasoning

**Your task:**
Find {num_solutions} real, existing solutions, algorithms, or methods in {domain} that address the shared relational structure. Look for published research, documented algorithms, or established methods.

**CRITICAL - Citation Accuracy Requirements:**
- Each solution MUST be based on a SPECIFIC paper, article, or documented method
- The cited source MUST directly describe the specific solution/algorithm/method you're reporting
- DO NOT cite general surveys or overview papers unless they specifically describe the method in detail
- DO NOT cite a source that only mentions the domain or related concepts - the source must describe THIS SPECIFIC SOLUTION
- Each solution should have DIFFERENT primary sources - if you're citing the same paper for multiple solutions, you're doing it wrong
- When in doubt, find MORE SPECIFIC papers that directly describe the method
- **CRITICAL:** You MUST provide the COMPLETE, EXACT TITLE of each paper/source you cite
- The paper title must match exactly what appears on the paper's webpage - do not truncate, abbreviate, or approximate titles

**Code Implementation Discovery:**
For each solution, try to find GitHub repositories with code implementations:
- Search for "[solution name] github implementation"
- Search for "[paper title] code"
- Check paper pages for "Code Availability" sections
- Look for official implementations or well-maintained research codebases
- If you cannot find verified GitHub repos, leave the github_repos array empty

For each solution found, return:
1. "title": Descriptive title of the solution/algorithm
2. "source_domain": "{domain}" (MUST be this exact domain name)
3. "description": 2-3 sentence explanation with specifics
4. "key_concepts": 3-5 core techniques/concepts
5. "relevance": How this solution addresses the shared relational structure and could transfer back to the biomedical domain
6. "sources": URLs or citations you found
7. "source_titles": EXACT titles of papers/articles at each source URL (must match order of sources array)
8. "github_repos": Array of GitHub repositories found (can be empty if none found)

**CRITICAL:** After completing your research, return ONLY the JSON array with NO additional text, explanation, or commentary. Do not write "Based on my research" or any other introduction. Start your response directly with the JSON array.

Format:
```json
[
  {{
    "title": "Solution name",
    "source_domain": "{{domain}}",
    "description": "Detailed explanation...",
    "key_concepts": ["concept1", "concept2", "concept3"],
    "relevance": "How this addresses the shared relational structure...",
    "sources": ["url1", "url2"],
    "source_titles": ["Exact Paper Title 1", "Exact Paper Title 2"],
    "github_repos": [
      {{
        "url": "https://github.com/owner/repo",
        "source": "paper"
      }}
    ]
  }}
]
```

Return ONLY the JSON array, nothing else.

\end{lstlisting}
\end{promptbox}

\subsection{Baseline Prompts}
\label{sec:baseline_prompts}

\begin{promptbox}[title={Cross-domain Domain Generation Prompt}]
\begin{lstlisting}[
    basicstyle=\ttfamily\scriptsize,
    breaklines=true,
    columns=fullflexible,
    keepspaces=true,
    frame=none,       % Turn off listing frame (box handles it)
    aboveskip=0pt,    % Remove extra gap at top
    belowskip=0pt     % Remove extra gap at bottom
]
You are analyzing a biomedical research problem to identify relevant domains for cross-domain transfer.

**Problem:**
{problem_text}

**Your task:**
Identify {num_domains} other domains (e.g., "computer_science", "logistics") that have relevant methods for addressing this problem.

**Example format:**
```json
["computer_science", "logistics"]
```

Return ONLY the JSON array, no other text.

\end{lstlisting}
\end{promptbox}

\begin{promptbox}[title={Cross-domain Search Prompt}]
\begin{lstlisting}[
    basicstyle=\ttfamily\scriptsize,
    breaklines=true,
    columns=fullflexible,
    keepspaces=true,
    frame=none,       % Turn off listing frame (box handles it)
    aboveskip=0pt,    % Remove extra gap at top
    belowskip=0pt     % Remove extra gap at bottom
]
You are finding solutions from {domain} to apply to a biomedical problem.

**Biomedical Problem:**
{problem_text}

**Your task:**
Find {num_solutions} real, existing solutions, algorithms, or methods from {domain} that could address this problem. Look for published research, documented algorithms, or established methods.

**OUTPUT CONCISENESS:**
- Keep descriptions to 2-3 sentences, focus on core algorithm/method only
- Avoid verbose background - provide only technical essentials

**CRITICAL - Citation Accuracy Requirements:**
- Each solution MUST be based on a SPECIFIC paper, article, or documented method
- The cited source MUST directly describe the specific solution/algorithm/method you're reporting
- DO NOT cite general surveys or overview papers unless they specifically describe the method in detail
- DO NOT cite a source that only mentions the domain or related concepts - the source must describe THIS SPECIFIC SOLUTION
- Each solution should have DIFFERENT primary sources - if you're citing the same paper for multiple solutions, you're doing it wrong
- When in doubt, find MORE SPECIFIC papers that directly describe the method
- **CRITICAL:** You MUST provide the COMPLETE, EXACT TITLE of each paper/source you cite
- The paper title must match exactly what appears on the paper's webpage - do not truncate, abbreviate, or approximate titles

**Code Implementation Discovery:**
For each solution, try to find GitHub repositories with code implementations:
- Search for "[solution name] github implementation"
- Search for "[paper title] code"
- Check paper pages for "Code Availability" sections
- Look for official implementations or well-maintained research codebases
- If you cannot find verified GitHub repos, leave the github_repos array empty

For each solution found, return:
1. "title": Descriptive title of the solution/algorithm
2. "source_domain": "{domain}" (MUST be this exact domain name)
3. "description": 2-3 sentence explanation with specifics
4. "key_concepts": 3-5 core techniques/concepts
5. "relevance": Explain how this solution could address the biomedical problem
6. "sources": URLs or citations you found
7. "source_titles": EXACT titles of papers/articles at each source URL (must match order of sources array)
8. "github_repos": Array of GitHub repositories found (can be empty if none found)

**CRITICAL:** After completing your research, return ONLY the JSON array with NO additional text, explanation, or commentary. Do not write "Based on my research" or any other introduction. Start your response directly with the JSON array.

Format:
```json
[
  {{
    "title": "Solution name",
    "source_domain": "{{domain}}",
    "description": "Detailed explanation...",
    "key_concepts": ["concept1", "concept2", "concept3"],
    "relevance": "How this addresses the biomedical problem...",
    "sources": ["url1", "url2"],
    "source_titles": ["Exact Paper Title 1", "Exact Paper Title 2"],
    "github_repos": [
      {{
        "url": "https://github.com/owner/repo",
        "source": "paper"
      }}
    ]
  }}
]
```

Return ONLY the JSON array, nothing else.

\end{lstlisting}
\end{promptbox}

\begin{promptbox}[title={No-domain Search Prompt}]
\begin{lstlisting}[
    basicstyle=\ttfamily\scriptsize,
    breaklines=true,
    columns=fullflexible,
    keepspaces=true,
    frame=none,       % Turn off listing frame (box handles it)
    aboveskip=0pt,    % Remove extra gap at top
    belowskip=0pt     % Remove extra gap at bottom
]
You are analyzing a biomedical research problem to find solutions.

**Problem:**
{problem_text}

**Your task:**
Find {num_solutions} real, existing solutions, algorithms, or methods that could address this problem. Look for published research, documented algorithms, or established methods.

**OUTPUT CONCISENESS:**
- Keep descriptions to 2-3 sentences, focus on core algorithm/method only
- Avoid verbose background - provide only technical essentials

**CRITICAL - Citation Accuracy Requirements:**
- Each solution MUST be based on a SPECIFIC paper, article, or documented method
- The cited source MUST directly describe the specific solution/algorithm/method you're reporting
- DO NOT cite general surveys or overview papers unless they specifically describe the method in detail
- DO NOT cite a source that only mentions the domain or related concepts - the source must describe THIS SPECIFIC SOLUTION
- Each solution should have DIFFERENT primary sources - if you're citing the same paper for multiple solutions, you're doing it wrong
- When in doubt, find MORE SPECIFIC papers that directly describe the method
- **CRITICAL:** You MUST provide the COMPLETE, EXACT TITLE of each paper/source you cite
- The paper title must match exactly what appears on the paper's webpage - do not truncate, abbreviate, or approximate titles

**Code Implementation Discovery:**
For each solution, try to find GitHub repositories with code implementations:
- Search for "[solution name] github implementation"
- Search for "[paper title] code"
- Check paper pages for "Code Availability" sections
- Look for official implementations or well-maintained research codebases
- If you cannot find verified GitHub repos, leave the github_repos array empty

For each solution found, return:
1. "title": Descriptive title of the solution/algorithm
2. "source_domain": A single domain name where this solution comes from (e.g., "computer_science", "logistics"). Do not combine multiple domains with slashes.
3. "description": 2-3 sentence explanation with specifics
4. "key_concepts": 3-5 core techniques/concepts
5. "relevance": Explain how this solution could address the biomedical problem
6. "sources": URLs or citations you found
7. "source_titles": EXACT titles of papers/articles at each source URL (must match order of sources array)
8. "github_repos": Array of GitHub repositories found (can be empty if none found)

**Example format:**
```json
[
  {{
    "title": "Solution name",
    "source_domain": "Domain name",
    "description": "Detailed explanation...",
    "key_concepts": ["concept1", "concept2", "concept3"],
    "relevance": "How this addresses the biomedical problem...",
    "sources": ["url1", "url2"],
    "source_titles": ["Exact Paper Title 1", "Exact Paper Title 2"],
    "github_repos": [
      {{
        "url": "https://github.com/owner/repo",
        "source": "paper"
      }}
    ]
  }}
]
```

Return ONLY the JSON array, no other text.

\end{lstlisting}
\end{promptbox}

\subsection{Query Generation Prompt}
\label{sec:query_generation_prompt}

\begin{promptbox}[title={Query Generation Prompt}]
\begin{lstlisting}[
    basicstyle=\ttfamily\scriptsize,
    breaklines=true,
    columns=fullflexible,
    keepspaces=true,
    frame=none,       % Turn off listing frame (box handles it)
    aboveskip=0pt,    % Remove extra gap at top
    belowskip=0pt     % Remove extra gap at bottom
]
Generate a short Semantic Scholar search query (3-4 terms only).

Solution methodology: {key_concepts}
Problem being solved: {problem_summary}

Return exactly 3-4 search terms on one line. Combine the most specific algorithm/technique name with a term describing the problem area.

Output only the terms, nothing else.

\end{lstlisting}
\end{promptbox}

\subsection{Rewritten Solution Transfer Prompt}
\label{sec:rewritten_solution_transfer_prompt}

\begin{promptbox}[title={Rewritten Solution Transfer Prompt}]
\begin{lstlisting}[
    basicstyle=\ttfamily\scriptsize,
    breaklines=true,
    columns=fullflexible,
    keepspaces=true,
    frame=none,       % Turn off listing frame (box handles it)
    aboveskip=0pt,    % Remove extra gap at top
    belowskip=0pt     % Remove extra gap at bottom
]
Rewrite a research paper title and abstract to emphasize application to a specific domain.

Original Title: {title}
Original Abstract: {description}
Key Technical Concepts: {key_concepts}
Target Application Domain: {problem_summary}

Your task:
1. Rewrite the TITLE to combine the methodology with the application (under 15 words)
   - Use the ACTUAL TECHNICAL METHODOLOGY from the key concepts, NOT any brand/algorithm names
   - Focus on the underlying technical approach (e.g., "graph neural networks", "matrix factorization") rather than named methods
   - MUST show how it's applied to the target domain
   - Example format: "[Technical Methodology] for [Target Domain Application]"

2. Rewrite the ABSTRACT to highlight practical application (~150-200 words)
   - Start with what problem in the target domain this method solves
   - Describe the technical methodology using the key concepts (avoid brand/algorithm names)
   - Explain how this technical approach addresses that specific problem
   - Focus on domain-specific application, not general theory

Return in this format:
```json
{{
  "title": "rewritten title combining methodology and application domain",
  "abstract": "rewritten abstract focusing on practical application to the specific problem domain"
}}
```

Return ONLY the JSON object, no other text.

\end{lstlisting}
\end{promptbox}

\subsection{Novelty Judge Prompts}
\label{sec:novelty_judge_prompts}

\begin{promptbox}[title={Stratified Novelty Judge Prompt}]
\begin{lstlisting}[
    basicstyle=\ttfamily\scriptsize,
    breaklines=true,
    columns=fullflexible,
    keepspaces=true,
    frame=none,       % Turn off listing frame (box handles it)
    aboveskip=0pt,    % Remove extra gap at top
    belowskip=0pt     % Remove extra gap at bottom
]
Proposed solution: {title}
Description: {description}
Key methodology being transferred: {key_concepts}
Target biomedical problem: {problem_summary}

Papers found that may describe similar work:
{papers_text}

Your task: Determine if the SPECIFIC METHODOLOGY from the proposed solution has already been applied to THIS SPECIFIC BIOMEDICAL DOMAIN.

CRITICAL: The novelty we care about is whether this methodology has been applied to THIS biomedical domain before - NOT whether the methodology exists in general. Finding papers that use the same method in other domains (climate, finance, etc.) does NOT reduce novelty - that's exactly what analogous reasoning is about!

Rate methodology overlap (0-10) - BE STRICT:

0-1: REVOLUTIONARY - This exact methodology has NEVER been explored in this biomedical domain
     - Zero papers found using this approach or related variants
     - This would be a paradigm-shifting first application

2-3: HIGHLY NOVEL - The specific methodology has not been applied, but some distant conceptual relatives exist
     - No papers use this methodology in this domain
     - Found papers address the biomedical problem with completely different approaches
     - Any methodological similarities are superficial (e.g., both use graphs, but different algorithms entirely)

4-6: MODERATE NOVELTY - Related methodologies exist but this specific approach has not been used
     - Papers use similar problem-solving paradigms but different specific techniques
     - Same family of methods (e.g., both use causality inference) but different algorithms or formulations
     - The core algorithmic innovation has not been applied, even if the general approach has been explored

7-8: LIMITED NOVELTY - Very similar methodologies already exist in this domain
     - Most key algorithmic components are already present in existing work
     - This would be a minor variation or incremental extension
     - The novelty is primarily in implementation details or parameter choices

9-10: NOT NOVEL - This exact methodology has been applied to this exact problem
      - Found papers describe essentially the same approach
      - Direct prior art exists with the same algorithmic approach
      - This is a re-discovery of existing work

Return ONLY valid JSON:
{{"methodology_overlap": N, "novelty_score": N, "assessment": "brief explanation - be explicit about whether the SPECIFIC methodology has been applied to this domain or not"}}

Where novelty_score = 10 - methodology_overlap (higher = more novel, range is now 0-10)

\end{lstlisting}
\end{promptbox}

\begin{promptbox}[title={Binary Novelty Judge Prompt}]
\begin{lstlisting}[
    basicstyle=\ttfamily\scriptsize,
    breaklines=true,
    columns=fullflexible,
    keepspaces=true,
    frame=none,       % Turn off listing frame (box handles it)
    aboveskip=0pt,    % Remove extra gap at top
    belowskip=0pt     % Remove extra gap at bottom
]
Proposed solution: {title}
Description: {description}
Key methodology being transferred: {key_concepts}
Target biomedical problem: {problem_summary}

Papers found that may describe similar work:
{papers_text}

Your task: Determine if the SPECIFIC METHODOLOGY from the proposed solution has already been applied to THIS SPECIFIC BIOMEDICAL DOMAIN.

CRITICAL: The novelty we care about is whether this methodology has been applied to THIS biomedical domain before - NOT whether the methodology exists in general. Finding papers that use the same method in other domains (climate, finance, etc.) does NOT reduce novelty - that's exactly what analogous reasoning is about!

Make a binary determination:

NOVEL:
- The specific methodology has NOT been applied to this biomedical domain
- No papers found using this approach or closely related variants in this domain
- Papers in this biomedical domain use fundamentally different approaches
- This represents a genuinely new application of the methodology to this domain

NOT NOVEL:
- The specific methodology HAS been applied to this biomedical domain
- Found papers that use this methodology or very similar approaches in this domain
- The core algorithmic approach already exists in the literature for this domain
- This would be incremental or duplicative work

Return ONLY valid JSON:
{{"is_novel": true/false, "assessment": "brief explanation - be explicit about whether the SPECIFIC methodology has been applied to this domain or not"}}

\end{lstlisting}
\end{promptbox}

\subsection{Analogy Judge Prompt}
\label{sec:analogy_judge_prompts}

\begin{promptbox}[title={Analogy Judge Prompt}]
\begin{lstlisting}[
    basicstyle=\ttfamily\scriptsize,
    breaklines=true,
    columns=fullflexible,
    keepspaces=true,
    frame=none,       % Turn off listing frame (box handles it)
    aboveskip=0pt,    % Remove extra gap at top
    belowskip=0pt     % Remove extra gap at bottom
]
You are an expert judge evaluating the quality of an analogy proposed to solve a scientific problem.

# Problem

**Problem Statement**: {problem}

# Analogy to Evaluate

**Problem Domain**: {source_domain}
**Analogous Domain**: {target_domain}

**Object Mappings**:
{object_mappings}

**Shared Relations**: {shared_relations}

# Evaluation Task

Score this analogy on 6 dimensions using a 0-10 scale.

**Structural Depth (0-10)**: How well-defined and meaningful are the object mappings?
- 0: Vague or superficial mappings with little explanatory power
- 10: Exceptional mappings with deep insight into structural correspondence

**Domain Distance (0-10)**: How far apart are the problem domain [{source_domain}] and analogous domain [{target_domain}]?
- 0: Same or closely related domains
- 10: Highly disparate domains with no obvious overlap

**Applicability (0-10)**: How well does this analogy enable knowledge transfer to solve the problem?
- 0: Analogy doesn't help solve the problem or is misleading
- 10: Transformative insight that directly enables solution

**Novelty (0-10)**: How original and insightful is this analogy?
- 0: Very common or obvious analogy
- 10: Groundbreaking insight, highly innovative

**Unexpectedness (0-10)**: How counter-intuitive or surprising is this analogy?
- 0: Obvious/expected connection that any expert would make
- 10: Highly surprising connection that requires significant conceptual leap

**Non-Obviousness (0-10)**: How unlikely is it that a domain expert would make this connection?
- 0: Standard analogy in the field's literature
- 10: Requires knowledge synthesis across disparate fields unlikely to intersect

# Output Format

Return ONLY valid JSON:

```json
{{
  "structural_depth": {{
    "score": <0-10>,
    "explanation": "<2-3 sentence justification>"
  }},
  "domain_distance": {{
    "score": <0-10>,
    "explanation": "<2-3 sentence justification>"
  }},
  "applicability": {{
    "score": <0-10>,
    "explanation": "<2-3 sentence justification>"
  }},
  "novelty": {{
    "score": <0-10>,
    "explanation": "<2-3 sentence justification>"
  }},
  "unexpectedness": {{
    "score": <0-10>,
    "explanation": "<2-3 sentence justification>"
  }},
  "non_obviousness": {{
    "score": <0-10>,
    "explanation": "<2-3 sentence justification>"
  }},
  "overall_assessment": "<2-3 sentence summary of the analogy's quality>"
}}
```

Return ONLY the JSON object, with no additional text.

\end{lstlisting}
\end{promptbox}

\subsection{Analogy Extraction Prompts}
\label{sec:analogy_extraction_prompts}

\begin{promptbox}[title={Baseline Analogy Extraction Prompt}]
\begin{lstlisting}[
    basicstyle=\ttfamily\scriptsize,
    breaklines=true,
    columns=fullflexible,
    keepspaces=true,
    frame=none,       % Turn off listing frame (box handles it)
    aboveskip=0pt,    % Remove extra gap at top
    belowskip=0pt     % Remove extra gap at bottom
]
You are analyzing a baseline LLM solution to extract its implicit analogical structure (if any).

**Problem Domain**: {source_domain}
**Solution Domain**: {target_domain}

**Problem:**
{problem}

**Baseline Solution:**
- Title: {solution_title}
- Description: {description}
- Key Concepts: {key_concepts}
- Relevance: {relevance}

**Your task:**
Extract the analogical structure if this solution represents cross-domain analogical reasoning.

**IMPORTANT:** If {source_domain} == {target_domain} (same domain), return empty "object_mappings": [] and set "shared_relations" to "Same-domain solution".

Otherwise, extract mappings showing how concepts from {target_domain} are used to solve the {source_domain} problem through explicit analogical mappings.

Look for explicit analogical language:
- "maps to", "corresponds to", "represents", "where X is Y"
- "nodes represent populations", "edges represent connections"
- "treating X as Y", "modeling X using Y"

**Return a JSON object with:**

1. "target_domain": The solution domain name (e.g., "{target_domain}")
2. "object_mappings": Array of problem-to-solution mappings with rationale (from {source_domain} to {target_domain})
3. "shared_relations": The relational structure preserved across domains

**Map objects by FUNCTION, not surface similarity.** "Delivers payload" is a good mapping basis; "is liquid" is not.

**Example format:**
```json
{{
  "target_domain": "domain_name",
  "object_mappings": [
    {{"source": "problem_object_A", "target": "solution_object_A", "mapping_rationale": "why these map functionally"}},
    {{"source": "problem_object_B", "target": "solution_object_B", "mapping_rationale": "why these map functionally"}}
  ],
  "shared_relations": "the relational structure preserved across domains"
}}
```

**Guidelines:**
- Extract 4-8 key object mappings if they are explicitly present (focus on functional mappings)
- Map by FUNCTION: what objects DO, not what they ARE
- **Direction**: source = problem domain ({source_domain}) object, target = solution domain ({target_domain}) object
- Mapping rationale should explain FUNCTIONAL similarity and quote/paraphrase the solution's language
- Only include mappings explicitly stated or strongly implied in the description/relevance
- If domains are the same, return empty mappings (all other cross-domain solutions should have mappings extracted)

Return ONLY the JSON object, no other text.

\end{lstlisting}
\end{promptbox}

\begin{promptbox}[title={Ground-Truth Analogy Extraction Prompt}]
\begin{lstlisting}[
    basicstyle=\ttfamily\scriptsize,
    breaklines=true,
    columns=fullflexible,
    keepspaces=true,
    frame=none,       % Turn off listing frame (box handles it)
    aboveskip=0pt,    % Remove extra gap at top
    belowskip=0pt     % Remove extra gap at bottom
]
You are analyzing a research paper that discovered an analogy between two domains.

**Problem Domain**: {source_domain}
**Analogous Domain**: {target_domain}
**Method Name**: {method_name}

**Analogy Description:**
{analogy_description}

**Concrete Example:**
{concrete_example}

**Your task:**
Extract the analogy's explicit object-to-object mappings from the paper's description.

IMPORTANT: Map from the PROBLEM domain ({source_domain}) to the ANALOGOUS domain ({target_domain}), matching the direction used in analogous reasoning workflows.

**Return a JSON object with:**

1. "target_domain": The analogous domain name (e.g., "{target_domain}")
2. "object_mappings": Array of problem-to-analogy mappings with rationale (from {source_domain} to {target_domain})
3. "shared_relations": The relational structure preserved across domains

**Map objects by FUNCTION, not surface similarity.** "Delivers payload" is a good mapping basis; "is liquid" is not.

**Example format:**
```json
{{
  "target_domain": "domain_name",
  "object_mappings": [
    {{"source": "problem_object_A", "target": "analogous_object_A", "mapping_rationale": "why these map functionally"}},
    {{"source": "problem_object_B", "target": "analogous_object_B", "mapping_rationale": "why these map functionally"}}
  ],
  "shared_relations": "the relational structure preserved across domains"
}}
```

**Guidelines:**
- Extract 4-8 key object mappings (focus on functional mappings)
- Map by FUNCTION: what objects DO, not what they ARE
- **Direction**: source = problem domain ({source_domain}) object, target = analogous domain ({target_domain}) object
- Mapping rationale should explain FUNCTIONAL similarity
- If the paper mentions surface similarities, reframe as functional mappings
- Extract from both analogy_description and concrete_example

Return ONLY the JSON object, no other text.

\end{lstlisting}
\end{promptbox}

\subsection{AR Dataset Prompts}
\label{sec:ar_dataset_prompts}

\begin{promptbox}[title={AR Dataset Discover Papers Prompt}]
\begin{lstlisting}[
    basicstyle=\ttfamily\scriptsize,
    breaklines=true,
    columns=fullflexible,
    keepspaces=true,
    frame=none,       % Turn off listing frame (box handles it)
    aboveskip=0pt,    % Remove extra gap at top
    belowskip=0pt     % Remove extra gap at bottom
]
Find up to {target_count} research papers that USE analogical reasoning as a METHOD to develop novel solutions, algorithms, or breakthroughs by transferring knowledge from {base_domain} to {target_domain}. Return as many as you can find (up to {target_count}), even if fewer than {target_count} exist.

CRITICAL REQUIREMENTS:
- **SOURCE DOMAIN**: Must draw from {base_domain}
- **TARGET DOMAIN**: Must apply to {target_domain}
- Find papers that either:
  1. PRESENT the original creation of a {target_domain} method/solution by applying an analogy from {base_domain}, OR
  2. APPLY an existing {base_domain} method to solve a problem in {target_domain}
- The paper should introduce a novel contribution or application (e.g., a new algorithm, technique, scientific breakthrough, or application)
- The analogy from {base_domain} should be the KEY INSIGHT that enabled the {target_domain} solution or application

DO NOT FIND:
- Papers ABOUT analogical reasoning (cognitive science, theoretical frameworks)
- Historical analyses of how someone used analogies
- Computational models OF analogical reasoning
- Survey papers, reviews, or book reviews that only describe methods without presenting applications
- Papers that only study or analyze analogical thinking
- Papers that merely optimize or incrementally improve existing {target_domain} methods without drawing on {base_domain} concepts

EXAMPLES OF WHAT TO FIND:
- Papers that introduce a new {target_domain} method/technique by applying a concept from {base_domain}
- Papers that solve a {target_domain} problem by transferring principles from {base_domain}
- Papers where {base_domain} insights led to {target_domain} breakthrough discoveries or innovations
- Papers that create novel {target_domain} approaches by drawing structural analogies from {base_domain}
- Papers that apply {base_domain} methods to solve {target_domain} problems

Focus on papers where the analogy from {base_domain} to {target_domain} is central to the contribution.

For each paper, provide:
1. The exact paper title
2. A DOI link (https://doi.org/...) or arXiv URL (https://arxiv.org/abs/...)
3. A brief description of what analogy is used (1 sentence)

IMPORTANT: Format your response as a valid JSON array like this example:
[
  {{"title": "Paper Title 1", "url": "https://doi.org/10.1234/example", "analogy_description": "Brief description of analogy used"}},
  {{"title": "Paper Title 2", "url": "https://arxiv.org/abs/1234.5678", "analogy_description": "Brief description of analogy used"}}
]
(Note: Use single braces in your actual response)

Return ONLY the JSON array, no other text.

\end{lstlisting}
\end{promptbox}

\begin{promptbox}[title={AR Dataset Extract Analogy Prompt}]
\begin{lstlisting}[
    basicstyle=\ttfamily\scriptsize,
    breaklines=true,
    columns=fullflexible,
    keepspaces=true,
    frame=none,       % Turn off listing frame (box handles it)
    aboveskip=0pt,    % Remove extra gap at top
    belowskip=0pt     % Remove extra gap at bottom
]
You are analyzing a research paper to determine if it USES analogical reasoning as a method to develop a novel solution, algorithm, or breakthrough.

Paper Title: {title}
Authors: {authors}
Year: {year}
Abstract: {abstract}

IMPORTANT: A paper USES analogical reasoning if the core method/solution it presents or applies was created by drawing an analogy from a different domain or subfield.

This includes:
1. **Original papers**: The paper PRESENTS the first creation of a method by applying an analogy from another field or subfield
2. **Application papers**: The paper APPLIES a method that was originally created through cross-domain analogical reasoning (even if this specific paper doesn't present the original creation)
3. **Empirical/evaluation papers**: The paper evaluates, tests, or improves a method that was created through analogical reasoning
4. **Review/commentary papers**: The paper DESCRIBES or discusses a method that was created through cross-domain analogical reasoning, even if it doesn't present new applications or evaluations

**What makes domains "different"?**
Two domains are sufficiently different if they have distinct:
- **Objects of study**: What phenomena, systems, or entities are being investigated?
- **Methodologies**: How is knowledge gained and validated in each domain?
- **Research traditions**: Do they have different historical origins or primary research communities?

**Important**: The existence of shared abstract concepts or similar terminology between two domains does NOT make them the same field. Domains can share conceptual vocabulary while still being sufficiently different in their objects of study, methods, and research communities. Interdisciplinary overlap or collaboration does NOT disqualify a transfer as analogical reasoning.

The paper does NOT qualify if:
- It is ABOUT analogical reasoning as a topic (cognitive science, historical analysis, computational modeling OF analogical reasoning itself)
- The work merely optimizes or incrementally improves an existing method without drawing on concepts from another area

The paper DOES qualify if:
- The base domain (source of inspiration) and target domain (where solution is applied) study different types of phenomena or have distinct primary research questions
- The method transfers mechanisms, principles, or structures from one domain to solve problems in another domain
- The paper presents concrete use, application, evaluation, improvement, or description of such a cross-domain method

If the paper does NOT use cross-domain analogical reasoning (neither presents, applies, evaluates, nor describes such a method), start your "problem" field with: "This paper does not use analogical reasoning to solve a problem."

If the paper DOES use analogical reasoning, extract the following information:

1. **Problem**: What is the core problem being solved by the analogical reasoning method discussed in this paper? For review/commentary papers, describe the problem that the original method addressed. Be specific and concise (1-2 sentences).

2. **Method Name**: The name of the method, algorithm, technique, or approach that uses the analogical reasoning. If the paper doesn't give it a specific name, create a brief descriptive name (e.g., "ecological competition model for cancer therapy", "PageRank for metabolic networks"). Keep it concise (3-8 words).

3. **Concrete Example**: Provide one specific, concrete example from the paper showing how the analogy is applied. This should be a tangible instance, not an abstract description. Include specific entities, mechanisms, or mappings (e.g., "The paper models tumor cell populations as competing species, where chemotherapy acts as a predator that shifts competitive advantages between drug-sensitive and drug-resistant cells"). (2-3 sentences)

4. **Base Domain**: The base domain name in simple lowercase format with underscores between words. Keep it concise (1-3 words maximum). This is the source domain from which the analogy is drawn.

5. **Target Domain**: The target domain name in simple lowercase format with underscores between words. Keep it concise (1-3 words maximum). This is the domain where the analogy is applied to solve the problem.

6. **Base Domain Justification**: What is the source domain from which the analogy is drawn? This is the domain that provides the analogy or inspiration. Name the field/domain clearly with full context and description.

7. **Target Domain Justification**: What is the target domain where the analogy is applied to solve the problem? This is the domain where the solution is being sought. Name the field/domain clearly with full context and description.

8. **Analogy Justification**: Explain why and how this analogy works. What are the structural or conceptual similarities that make this cross-domain transfer valid? Importantly, identify the specific MECHANISM or PRINCIPLE being transferred from the base domain and explain how it maps onto the target domain problem. (2-3 sentences)

9. **Is Original Paper**: Is this paper the ORIGINAL paper that first presented/created this analogical reasoning method, or is it a review/evaluation/application of an existing method? Answer true if this is the original, false if it's a later paper about the method.

   **IMPORTANT**: If the paper title mentions specific researchers' names (e.g., "Smith and Jones: ...") but those researchers are NOT the actual authors of this paper, this is almost certainly a retrospective/commentary paper about their original work, not the original paper itself. Mark as is_original_paper: false.

10. **Original Paper Info**: If this is NOT the original paper (is_original_paper = false), provide information about what the original paper would be (authors, approximate year, or description). If this IS the original paper, leave this empty.

11. **Domain Distance**: How different are the base and target domains? Rate as one of the following:
   - **"distant"**: Domains study fundamentally different phenomena with little to no traditional overlap (e.g., biology -> computer science, physics -> economics)
   - **"moderate"**: Domains have some conceptual overlap but distinct primary research questions and methodologies (e.g., machine learning -> robotics, psychology -> education)
   - **"close"**: Domains share significant methodological overlap, research communities, or are subfields of a common parent field (e.g., supervised learning -> unsupervised learning, organic chemistry -> biochemistry)

12. **Domain Distance Justification**: Briefly explain your domain distance rating. What makes these domains distant, moderate, or close? (1-2 sentences)

13. **Analogy Usage Depth**: How is the analogy being used in this paper? Rate as one of the following:
   - **"conceptual_motivation"**: Analogy provides inspiration, framing, or intuition but doesn't transfer concrete methods or tools
   - **"methodological_adaptation"**: Specific methods, tools, or approaches are adapted from the source domain
   - **"deep_structural_transfer"**: Core mechanisms, mathematical structures, or algorithmic principles are directly transferred and form the foundation of the solution

14. **Analogy Usage Justification**: Explain what specifically is being transferred (if anything) and how central it is to the paper's contribution. (2-3 sentences)

15. **Requires Structural Reasoning**: Could an LLM suggest this domain connection from keywords/semantic similarity alone, or does it require understanding the underlying mechanism?

Answer TRUE if discovering this connection requires structural/mechanistic insight (not obvious from surface features).
Answer FALSE if the domain connection is semantically obvious or well-known.

16. **Structural Reasoning Justification**: Briefly explain your answer. (1-2 sentences)

17. **Likely Well-Known Example**: Is this a famous/widely-cited cross-domain application that would appear frequently in academic literature and textbooks?

Answer TRUE if this is a classic, well-known example. Answer FALSE if obscure, recent, or novel.

18. **Well-Known Justification**: Briefly explain. (1 sentence)

Return your answer in the following JSON format:
{{
  "problem": "...",
  "method_name": "...",
  "concrete_example": "...",
  "base_domain": "...",
  "target_domain": "...",
  "base_domain_justification": "...",
  "target_domain_justification": "...",
  "analogy_justification": "...",
  "is_original_paper": true,
  "original_paper_info": "",
  "domain_distance": "distant",
  "domain_distance_justification": "...",
  "analogy_usage_depth": "deep_structural_transfer",
  "analogy_usage_justification": "...",
  "requires_structural_reasoning": true,
  "structural_reasoning_justification": "...",
  "likely_well_known_example": false,
  "well_known_justification": "..."
}}

Be specific, concise, and focus on the analogical reasoning aspect of the paper.

\end{lstlisting}
\end{promptbox}

\begin{promptbox}[title={AR Dataset Assess Analogy Difficulty Prompt}]
\begin{lstlisting}[
    basicstyle=\ttfamily\scriptsize,
    breaklines=true,
    columns=fullflexible,
    keepspaces=true,
    frame=none,       % Turn off listing frame (box handles it)
    aboveskip=0pt,    % Remove extra gap at top
    belowskip=0pt     % Remove extra gap at bottom
]
You are assessing the difficulty of analogical reasoning used in a research paper.

Paper: {title}

**Analogy Details:**
- Base Domain: {base_domain}
- Target Domain: {target_domain}
- Justification: {justification}

**Assessment Criteria:**

**EASY**: Well-known, textbook analogies or straightforward connections
- Examples: Standard applications everyone learns, obvious similarities, commonly taught methods

**MEDIUM**: Requires some creative insight but not deeply surprising
- Examples: Specific technical methods applied in new ways, connections that make sense once explained

**HARD**: Creative leap requiring deep insight, non-obvious connections
- Examples: Counter-intuitive mappings, obscure mechanisms, groundbreaking analogies, abstract structural similarities

Based on these criteria, assess the difficulty of this analogical reasoning.

Return your answer in the following JSON format:
{{
  "difficulty": "easy|medium|hard",
  "reasoning": "Brief explanation for your assessment (1-2 sentences)"
}}

\end{lstlisting}
\end{promptbox}

\begin{promptbox}[title={AR Dataset Rewrite Problem Prompt}]
\begin{lstlisting}[
    basicstyle=\ttfamily\scriptsize,
    breaklines=true,
    columns=fullflexible,
    keepspaces=true,
    frame=none,       % Turn off listing frame (box handles it)
    aboveskip=0pt,    % Remove extra gap at top
    belowskip=0pt     % Remove extra gap at bottom
]
Rewrite this problem to its simplest form - stating ONLY the goal, with NO hints about mechanisms or methods.

Original Problem:
{problem}

Solution domain to hide: {base_domain}
Target domain: {target_domain}

Rules:
1. State only what needs to be achieved
2. Remove all terminology from the solution domain: {base_domain}
3. Remove all "by...", "through...", "using..." phrases
4. Format: "How to [goal]"

Return only the rewritten problem:
\end{lstlisting}
\end{promptbox}

\section{Example Generations}
\label{sec:example_generations}
We report one example generation for each of the three settings (no-domain, cross-domain, AR) using the exact parameters used for our evaluation tasks. We utilize the research problem ``How to predict and characterize the collective behavior of T cell populations and their receptor recognition patterns across diverse immune repertoires'' from one of the papers in the AR Dataset.

\begin{promptbox}[title={No-domain Generation}]
\begin{lstlisting}[
    basicstyle=\ttfamily\scriptsize,
    breaklines=true,
    columns=fullflexible,
    keepspaces=true,
    frame=none,       % Turn off listing frame (box handles it)
    aboveskip=0pt,    % Remove extra gap at top
    belowskip=0pt     % Remove extra gap at bottom
]
============================================================
FOUND 1 SOLUTIONS
============================================================

1. GLIPH2: Grouping of Lymphocyte Interactions by Paratope Hotspots (computational_immunology)
   GLIPH2 is an algorithm that identifies shared specificity motifs in TCR sequences by clustering receptors with similar CDR3 sequences and analyzing local and global similarity patterns to predict antigen specificity. It uses probabilistic models to assess the significance of TCR sequence convergence across repertoires and incorporates HLA restriction information to improve specificity predictions.
   Concepts: TCR sequence clustering, CDR3 motif analysis, convergence probability modeling, HLA restriction integration, specificity group identification
   Sources (1):
    Identification of specificity groups in the T cell receptor repertoire
        https://www.nature.com/articles/s41587-020-0505-4
\end{lstlisting}
\end{promptbox}

\begin{promptbox}[title={Cross-domain Generation}]
\begin{lstlisting}[
    basicstyle=\ttfamily\scriptsize,
    breaklines=true,
    columns=fullflexible,
    keepspaces=true,
    frame=none,       % Turn off listing frame (box handles it)
    aboveskip=0pt,    % Remove extra gap at top
    belowskip=0pt     % Remove extra gap at bottom
]
Discovering 1 domains, then finding 1 solutions per domain...
Discovered domains: ecology
  [1/1] Finding solutions in ecology...

============================================================
FOUND 1 SOLUTIONS
============================================================

1. Neutral Theory Models for Predicting Community Abundance Distributions (ecology)
   Hubbell's unified neutral theory provides a framework for predicting species abundance distributions in ecological communities based on fundamental parameters of speciation, migration, and demographic stochasticity without requiring species-specific differences. The model uses sampling theory and maximum likelihood estimation to infer metacommunity biodiversity and migration rates from local community samples.
   Concepts: neutral theory, species abundance distribution, metacommunity dynamics, maximum likelihood estimation, demographic stochasticity
   Sources (2):
    A Unified Neutral Theory of Biodiversity
        https://www.science.org/doi/10.1126/science.286.5442.1146
    Neutral Theory and Relative Species Abundance in Ecology
        https://www.ncbi.nlm.nih.gov/pmc/articles/PMC1691315/
\end{lstlisting}
\end{promptbox}

\begin{promptbox}[title={AR Generation}]
\begin{lstlisting}[
    basicstyle=\ttfamily\scriptsize,
    breaklines=true,
    columns=fullflexible,
    keepspaces=true,
    frame=none,       % Turn off listing frame (box handles it)
    aboveskip=0pt,    % Remove extra gap at top
    belowskip=0pt     % Remove extra gap at bottom
]
============================================================
PROBLEM EXTRACTION
============================================================

Summary: The problem involves predicting and characterizing how diverse populations of T cells collectively recognize antigens through their receptors, requiring understanding of recognition patterns across variable immune repertoires.

Problem Objects:
  - T cell population: Diverse collection of recognition units that collectively respond to threats
  - T cell receptor (TCR): Variable recognition interface that binds to specific molecular patterns
  - Antigen/epitope: Target molecular pattern to be recognized and classified
  - Immune repertoire: Complete catalog of available recognition capabilities across individuals or contexts
  - Recognition pattern: Characteristic binding profile or signature that defines specificity
  - Collective behavior: Emergent response properties arising from population-level interactions and diversity

Core Relations:
  - Multiple diverse recognition units (TCRs) map to specific target patterns (antigens)
  - Recognition specificity varies across the population creating a distributed detection system
  - Individual recognition events aggregate to produce collective population-level responses
  - Repertoire diversity determines the breadth and reliability of pattern recognition
  - Predictive models must account for both individual specificity and population-level statistics

Analogies:

  [CYBERSECURITY] Intrusion Detection Systems and Signature-Based Threat Recognition
    Object Mappings:
      T cell population -> Distributed network of detection sensors/agents (Both represent diverse collections of monitoring units that collectively provide security/defense coverage)
      T cell receptor (TCR) -> Threat detection signature/pattern matcher (Both are variable recognition interfaces designed to identify specific patterns among vast possibilities)
      Antigen/epitope -> Malware signature or attack pattern (Both are specific target patterns that must be recognized and classified by the detection system)
      Immune repertoire -> Signature database or rule set library (Both represent the complete catalog of recognition capabilities available for threat detection)
      Recognition pattern -> Detection heuristic or matching rule (Both define specific criteria and profiles used to identify threats with certain specificity and sensitivity)
      Collective behavior -> Aggregate security posture from distributed monitoring (Both represent emergent system-level defense capabilities arising from multiple independent detection events)
    Shared Relations: Multiple diverse detection mechanisms scan for specific threat patterns; individual detection events aggregate to create system-wide security responses; repertoire diversity determines coverage breadth; predictive models must balance individual signature accuracy with population-level statistical behavior to anticipate novel threats and minimize false positives

Key Terms: T cell receptor repertoire, antigen recognition, collective immune response, epitope binding patterns, repertoire diversity, population-level prediction, receptor specificity, distributed recognition system
Target Domains: cybersecurity

============================================================
FOUND 1 SOLUTIONS
============================================================

1. PAYL: Anomalous Payload-based Intrusion Detection (cybersecurity)
   PAYL uses statistical models of normal network payload byte distributions to detect anomalies, computing a Mahalanobis distance measure from learned models of typical traffic patterns. The system builds multiple distinct models for different network services, creating a diverse detection repertoire where collective anomaly scores from distributed sensors determine intrusion likelihood, balancing individual signature sensitivity with population-level statistical behavior.
   Concepts: statistical payload modeling, Mahalanobis distance anomaly scoring, distributed model repertoire, byte frequency distribution analysis, aggregate anomaly detection
   Sources (1):
    PAYL: Anomalous Payload-based Intrusion Detection
        https://www.cs.columbia.edu/~angelos/Papers/2004/payl.pdf

\end{lstlisting}
\end{promptbox}

\section{Additional Plots}
\label{sec:additional_plots}
\begin{figure}[h]
\begin{center}
\includegraphics[scale=0.6]{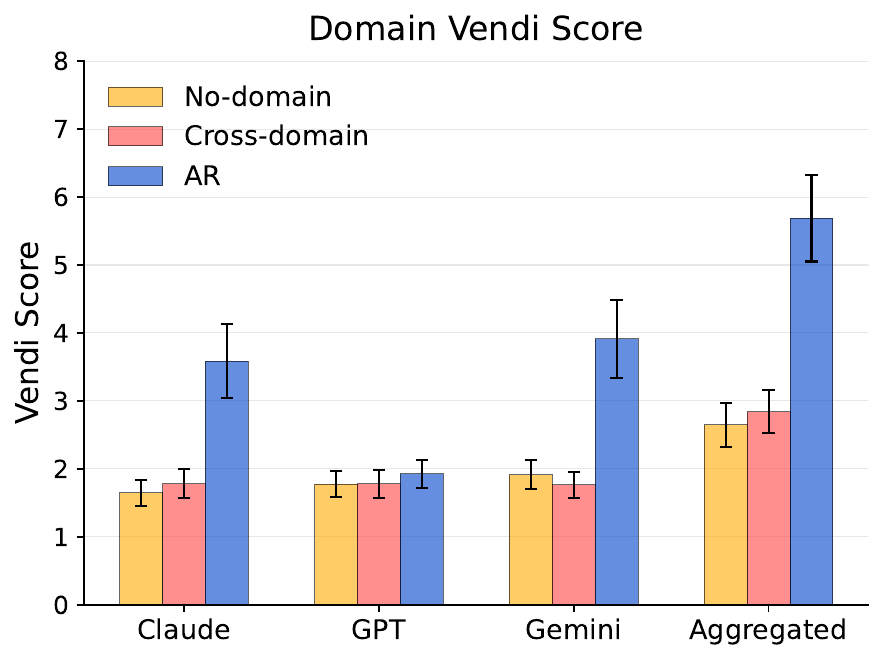}
\caption{Average domain Vendi Score with error bars of 95\% confidence intervals. Evaluated across 50 research problems with 50 solutions generated per problem. Results are shown for three settings (no-domain, cross-domain, AR) for Claude, GPT, Gemini, and an aggregated set of solutions from all three LLMs.}
\label{fig:domain_vendi_bar_chart}
\end{center}
\end{figure}

\begin{figure}[h]
\begin{center}
\includegraphics[scale=0.6]{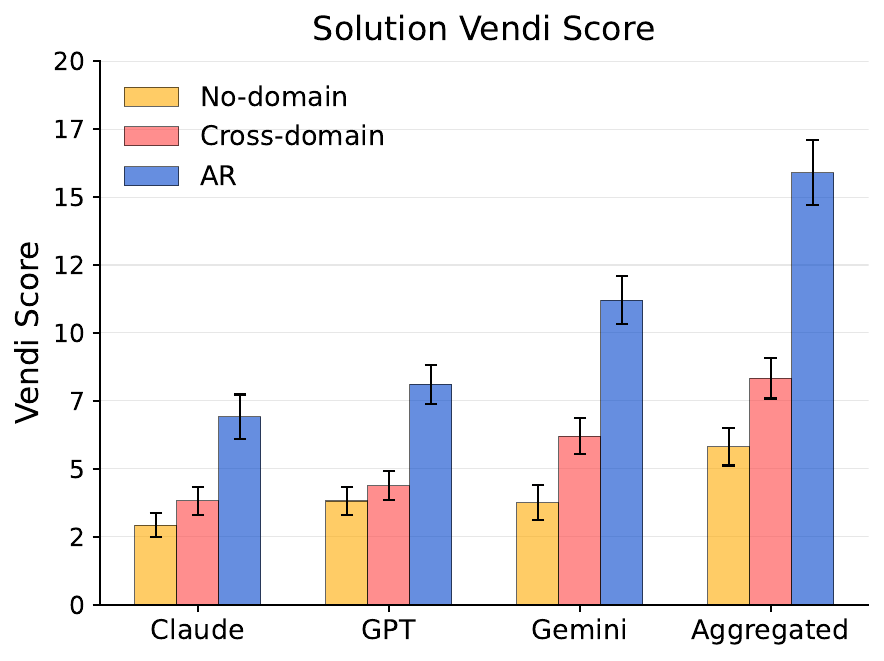}
\caption{Average solution Vendi Score with error bars of 95\% confidence intervals. Evaluated across 50 research problems with 50 solutions generated per problem. Results are shown for three settings (no-domain, cross-domain, AR) for Claude, GPT, Gemini, and an aggregated set of solutions from all three LLMs.} 
\label{fig:solution_vendi_bar_chart}
\end{center}
\end{figure}

\begin{figure}[h]
\begin{center}
\includegraphics[scale=0.6]{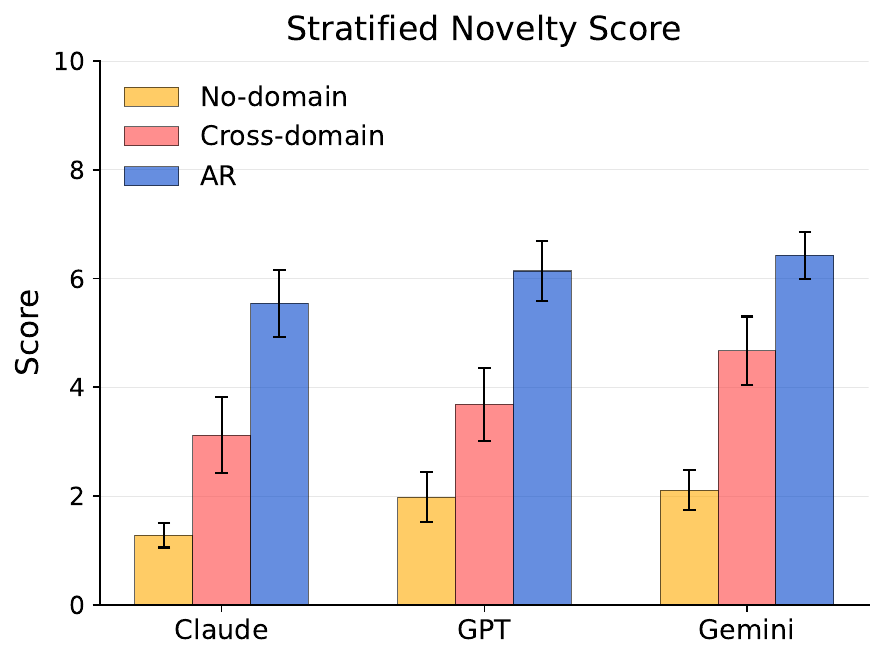}
\caption{Average of mean per-problem novelty scores with error bars of 95\% confidence intervals. Evaluated across 50 research problems with 5 solutions generated per problem. Results are shown for three settings (no-domain, cross-domain, AR) for three LLMs (Claude, GPT, Gemini) for the Stratified LLM-judge scoring prompt.}
\label{fig:stratified_novelty_bar_chart}
\end{center}
\end{figure}

\begin{figure}[h]
\begin{center}
\includegraphics[scale=0.6]{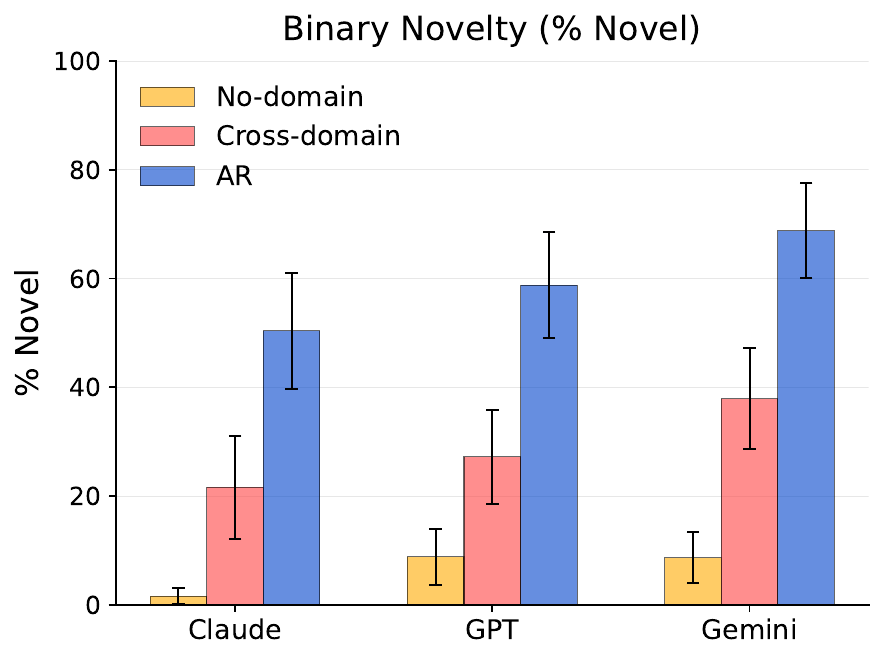}
\caption{Average of mean per-problem novelty scores with error bars of 95\% confidence intervals. Evaluated across 50 research problems with 5 solutions generated per problem. Results are shown for three settings (no-domain, cross-domain, AR) for three LLMs (Claude, GPT, Gemini) for the Binary LLM-judge scoring prompt.}
\label{fig:binary_novelty_bar_chart}
\end{center}
\end{figure}

\begin{figure}[h]
\begin{center}
\includegraphics[scale=0.5]{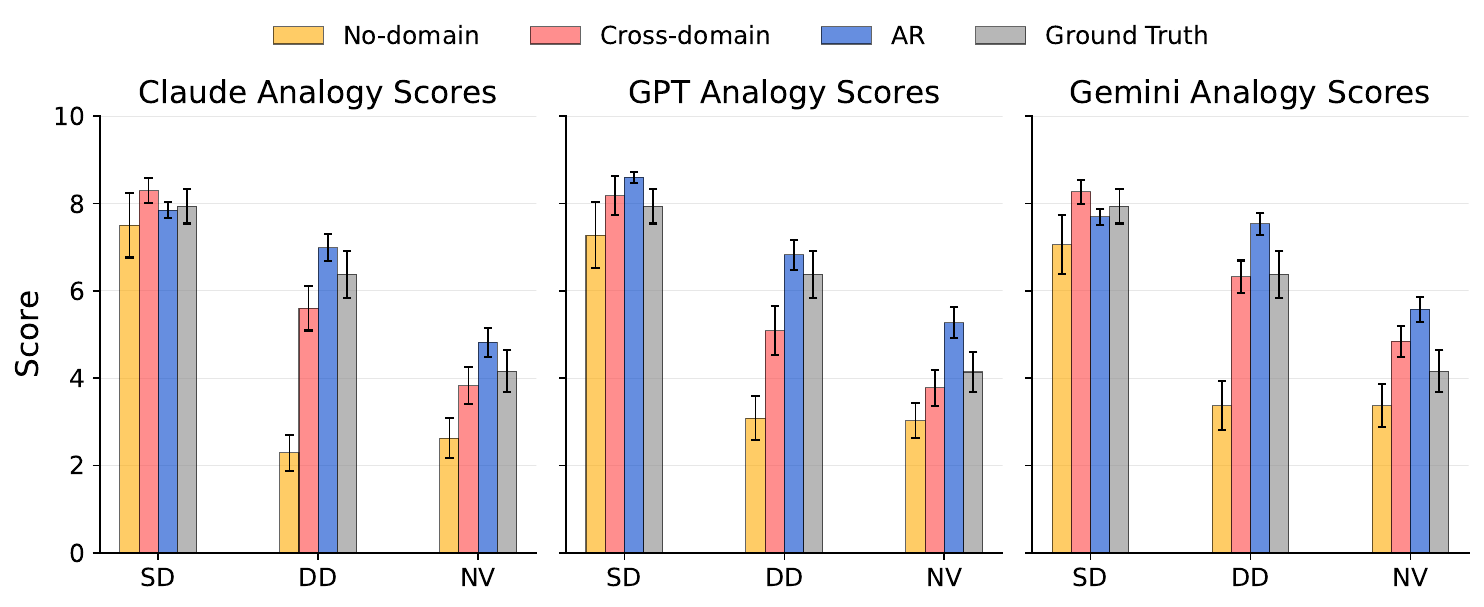}
\caption{Average of mean per-problem analogy scores with error bars of 95\% confidence intervals. Evaluated across 50 research problems with 5 solutions generated per problem. Results are shown for four settings (no-domain, cross-domain, AR, ground-truth) for three LLMs (Claude, GPT, Gemini) and assessed across three LLM-judge metrics (Structural Depth: SD, Domain Distance: DD, Novelty: NV).} 
\label{fig:analogy_results_all}
\end{center}
\end{figure}

We include additional figures including the domain Vendi Score (see Figure \ref{fig:domain_vendi_bar_chart}), solution Vendi Score (see Figure \ref{fig:solution_vendi_bar_chart}), solution novelty in bar chart format (see Figure \ref{fig:stratified_novelty_bar_chart}, Figure \ref{fig:binary_novelty_bar_chart}), and analogy quality scores across all LLMs (see Figure \ref{fig:analogy_results_all}).

We report the full diversity metrics across all LLMs and all settings in Table  \ref{table:diversity}. In addition to the domain Vendi Score and the solution Vendi Score, we report the number of unique domains and unique solutions generated by the various settings.

\begin{table}[h]
\centering
\setlength{\tabcolsep}{4pt}
\renewcommand{\arraystretch}{1.3}
\begin{tabular}{l c c c c c}
\toprule
{LLM} & {Setting} & \makecell[c]{{Unique} \\ {domains} \\ ($\uparrow$)} & \makecell[c]{{Domain} \\ {Vendi Score} \\ ($\uparrow$)} & \makecell[c]{{Unique} \\ {solutions} \\ ($\uparrow$)} & \makecell[c]{{Solution} \\ {Vendi Score} \\ ($\uparrow$)} \\
\midrule
 & No-domain & 3.42 & 1.64 & 25.44 & 2.93 \\
Claude & Cross-domain & 3.86 & 1.78 & 27.78 & 3.82 \\
 & AR & \textbf{10.30} & \textbf{3.58} & \textbf{39.24} & \textbf{6.92} \\
\midrule
 & No-domain & 3.34 & 1.77 & 39.42 & 3.81 \\
GPT & Cross-domain & 3.64 & 1.77 & 38.42 & 4.38 \\
 & AR & \textbf{4.50} & \textbf{1.92} & \textbf{47.90} & \textbf{8.10} \\
\midrule
 & No-domain & 4.12 & 1.91 & 21.82 & 3.75 \\
Gemini & Cross-domain & 3.18 & 1.76 & 32.96 & 6.20 \\
 & AR & \textbf{15.42} & \textbf{3.91} & \textbf{46.28} & \textbf{11.19} \\
\midrule
 & No-domain & 7.40 & 2.64 & 86.42 & 5.81 \\
Aggregated & Cross-domain & 7.90 & 2.84 & 98.96 & 8.34 \\
 & AR & \textbf{27.56} & \textbf{5.68} & \textbf{133.32} & \textbf{15.90} \\
\bottomrule
\end{tabular}
\caption{Average of mean per-problem diversity metrics. Evaluated across 50 research problems with 50 solutions generated per problem. Results are shown for three settings (no-domain, cross-domain, AR) for Claude, GPT, Gemini, and an aggregated set of solutions from all three LLMs.}
\label{table:diversity}
\end{table}

We also report the full analogy quality scores across all LLMs and all settings in Table \ref{table:analogy}. Though we discuss the 3 main LLM-judged metrics (Structural Depth, Domain Distance, Novelty) in the paper, we also measure 3 other metrics (Applicability, Unexpectedness, Non-Obviousness). Higher analogy scores on Applicability seem to correspond with lower Novelty scores, and Unexpectedness and Non-Obviousness exhibit the same trends as Novelty. We also report the number of Valid Analogies calculated using the criteria from Section \ref{sec:analogy_extraction}.

\begin{table}[h]
\centering
\setlength{\tabcolsep}{4pt}
\renewcommand{\arraystretch}{1.3}
\begin{tabular}{l c c c c c c c c}
\toprule
{LLM} & {Setting} & \makecell[c]{{SD} \\ ($\uparrow$)} & \makecell[c]{{DD} \\ ($\uparrow$)} & \makecell[c]{{AP} \\ ($\uparrow$)} & \makecell[c]{{NV} \\ ($\uparrow$)} & \makecell[c]{{UN} \\ ($\uparrow$)} & \makecell[c]{{NO} \\ ($\uparrow$)} & \makecell[c]{{Valid Analogies} \\ ($\uparrow$)} \\
\midrule
~ & No-domain & 7.51 & 2.29 & \textbf{8.76} & 2.62 & 1.85 & 1.47 & 93/250 \\
Claude & Cross-domain & \textbf{8.30} & 5.60 & 8.74 & 3.83 & 3.43 & 3.05 & 238/250 \\
~ & AR & 7.84 & \textbf{6.99} & 7.17 & \textbf{4.82} & \textbf{4.76} & \textbf{4.58} & \textbf{250/250} \\
~ & Ground Truth & 7.94 & 6.38 & 7.40 & 4.16 & 3.98 & 3.72 & N/A \\
\midrule
~ & No-domain & 7.28 & 3.08 & \textbf{8.75} & 3.03 & 2.13 & 1.90 & 104/250 \\
GPT & Cross-domain & 8.18 & 5.09 & 8.69 & 3.78 & 3.10 & 2.97 & 220/250 \\
~ & AR & \textbf{8.60} & \textbf{6.83} & 7.97 & \textbf{5.27} & \textbf{4.82} & \textbf{4.88} & \textbf{249/250} \\
~ & Ground Truth & 7.94 & 6.38 & 7.40 & 4.16 & 3.98 & 3.72 & N/A \\
\midrule
~ & No-domain & 7.06 & 3.37 & 8.08 & 3.37 & 2.57 & 2.39 & 108/250 \\
Gemini & Cross-domain & \textbf{8.27} & 6.32 & \textbf{8.46} & 4.85 & 4.45 & 4.22 & 243/250 \\
~ & AR & 7.69 & \textbf{7.53} & 6.86 & \textbf{5.58} & \textbf{5.73} & \textbf{5.67} & \textbf{248/250} \\
~ & Ground Truth & 7.94 & 6.38 & 7.40 & 4.16 & 3.98 & 3.72 & N/A \\
\bottomrule
\end{tabular}
\caption{Average of mean per-problem analogy scores. Evaluated across 50 research problems with 5 solutions generated per problem. Results are shown for four settings (no-domain, cross-domain, AR, ground-truth) for three LLMs (Claude, GPT, Gemini) and assessed across six LLM-judge metrics (Structural Depth: SD, Domain Distance: DD, Applicability: AP, Novelty: NV, Unexpectedness: UN, Non-Obviousness: NO). Valid Analogies denotes the number of valid analogies out of 250 generations.}
\label{table:analogy}
\end{table}

\section{Broader Impacts and Safety}
\label{sec:ethics_statement}
While analogical reasoning for open-ended solution generation is designed to overcome mode collapse and discover novel approaches to biomedical problems, the system could theoretically be exploited by bad actors to generate harmful biological agents or protocols. We emphasize the importance of careful deployment and rigorous safety reviews to prevent the implementation of such risks. In addition, our analogical reasoning system is intended to augment human researchers by discovering more creative approaches to solving research problems. As autonomous science research progresses, rigorous empirical evaluation of the feasibility and safety of these AI-generated solutions will remain of the utmost importance.


\end{document}